%% file: main.tex
\documentclass[10pt,twocolumn,letterpaper]{article}

\usepackage{cvpr}              % To produce the CAMERA-READY version
\input{preamble}
\definecolor{cvprblue}{rgb}{0.21,0.49,0.74}
\usepackage[pagebackref,breaklinks,colorlinks,allcolors=cvprblue]{hyperref}
\usepackage{booktabs}
\usepackage{multirow}
\usepackage{graphicx} % For \rotatebox
\usepackage{amsmath}
\usepackage{makecell}

\def\paperID{****} % *** Enter the Paper ID here
\def\confName{CVPR}
\def\confYear{2026}

\title{AVA-VLA: Improving Vision-Language-Action models with Active \\ Visual Attention}

\author{Lei Xiao\textsuperscript{1, $^*$}\;\;\; Jifeng Li\textsuperscript{1, \thanks{Equal contribution.}}\;\;\; Juntao Gao\textsuperscript{1,2}\;\;\; Feiyang Ye\textsuperscript{1, \thanks{Corresponding author.}}\;\;\; \\ Yan Jin\textsuperscript{1}\;\;\; Jingjing Qian\textsuperscript{3}\;\;\; Jing Zhang\textsuperscript{2} \;\;\; Yong Wu\textsuperscript{1} \;\;\; Xiaoyuan Yu\textsuperscript{1, $^\dagger$} \;\;\;\vspace{2mm}\\
\textsuperscript{1}LiAuto Inc.\;\;\; 
\textsuperscript{2}Beijing University of Technology \;\;\;  \textsuperscript{3}The Chinese University of Hong Kong, Shenzhen\\
}

\begin{document}
\maketitle
\input{sec/0_abstract}    
\input{sec/1_intro}
\input{sec/2_related_work}

\input{sec/3_method}

\input{sec/4_experiment}

\input{sec/5_conclusion}

% \input{sec/0_abstract}    
% \clearpage
% \input{sec/1_intro}
% \clearpage
% \input{sec/2_related_work}
% \clearpage
% \input{sec/3_method}
% \clearpage
% \input{sec/4_experiment}
% \clearpage
% \input{sec/5_conclusion}
% \clearpage

\section*{Acknowledgements}
This work is supported by Beijing Natural Science Foundation (No. L247025).

{
    \small
    \bibliographystyle{ieeenat_fullname}
    \bibliography{main}
}

\input{sec/X_suppl}

\end{document}

%% file: sec/0_abstract.tex
\begin{abstract}

Vision-Language-Action (VLA) models have shown remarkable progress in embodied tasks recently, but most methods process visual observations independently at each timestep. This history-agnostic design treats robot manipulation as a Markov Decision Process, even though real-world robotic control is inherently partially observable and requires reasoning over past interactions. To address this mismatch, we reformulate VLA policy learning from a Partially Observable Markov Decision Process perspective and propose AVA-VLA, a framework that conditions action generation on a recurrent state that serves as a neural approximation to the agent’s belief over task history. Built on this recurrent state, we introduce Active Visual Attention (AVA), which dynamically reweights visual tokens in the current observation to focus on regions most relevant given both the instruction and execution history. Extensive experiments show that AVA-VLA achieves state-of-the-art performance on standard robotic benchmarks, including LIBERO and CALVIN, and transfers effectively to real-world dual-arm manipulation tasks. These results demonstrate the effectiveness of temporally grounded active visual processing for improving VLA performance in robotic sequential decision-making. The project page is available at \url{https://liauto-dsr.github.io/AVA-VLA-Page}.

% in both simulation and the real world 
%Our analysis further validates that ATA effectively identifies xxx, underscoring the benefits of the active attention paradigm.

% Decision-State Modulation

% Intentional Priors

% Internal State Feedback

% History-as-Guidance

% Recurrent Latents

\end{abstract}

%% file: sec/1_intro.tex
\section{Introduction}
\label{sec:intro}
Recent advances in robotic manipulation have demonstrated impressive progress in training robot action policies that can act across diverse real-world tasks. One transformative paradigm is Vision-Language-Action (VLA) models \cite{rt1, rt2, bjorck2025gr00t, cheang2025gr, openvla, pi0, pi0-fast, li2024cogact}, which integrate visual perception, natural language understanding, and action generation within a unified neural architecture. These models, which are capable of instruction following and robotic action generation, exhibit strong understanding and generalization abilities after being fine-tuned for downstream scenarios.

\begin{figure}[tbp]
\centering
\includegraphics[width=1\linewidth]{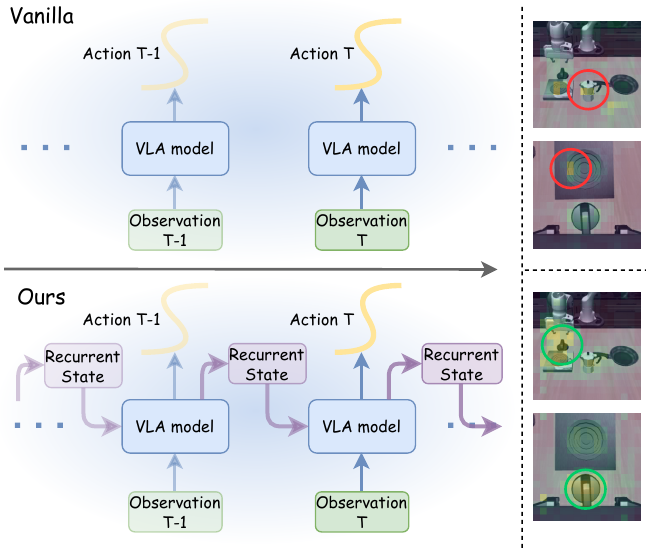}
\vskip -0.01in
\caption{(a) Visualized comparison of the proposed AVA-VLA framework and vanilla VLAs.  (b) Qualitative comparison of visual focus from two viewpoints while executing the task ``turn on the stove and put the moka pot on it." The vanilla OpenVLA-OFT~\cite{openvla-oft} baseline fails to locate the task-critical ``stove" switch, whereas AVA-VLA exhibits more stable focus by leveraging historical context.}
\label{fig:figure_1}
\vskip -0.17in
\end{figure}

To adopt the ability to understand diverse scenes, objects, and language instructions, most VLA models are built upon pretrained Vision-Language Models (VLMs) \cite{liu2023visual, chen2023pali, karamcheti2024prismatic}. Such models typically extend VLM architectures with modules such as action tokenization \citep{openvla, song2025accelerating} or specialized action experts \citep{tinyvla, li2024cogact} to enable action-oriented outputs. Based on this architectural inheritance, these VLA models typically process visual inputs as isolated temporal frames, treating each frame independently. This implicitly formulates robot manipulation as a Markov Decision Process (MDP) \cite{mees2022calvin, jiang2025irl}, where actions are generated from the current visual observation, assumed to represent the complete world state. In realistic robotic manipulation, however, the current visual frame is only a partial observation of the environment state. This full state includes unobservable dynamics across time, such as internal states and occluded information. By discarding the rich context from the past, this MDP-based approach is suboptimal for the dynamic sequential decision-making required in robotic manipulation.

This limitation of the MDP-based assumption has significant impacts on VLA models, particularly for the model's visual processing capabilities. VLA modeling is essentially a dynamic feedback control problem, where each preceding action directly alters the current visual input. However, by processing frames in isolation, the visual attention weights, guided by the static language instruction, are forced to re-evaluate the independent visual information from scratch at each decision step. Without global context, the model cannot effectively suppress temporally redundant information and focus on regions made important by past actions. As a result, the visual system remains passive rather than active. In fact, the inability to anticipate perceptual intent a priori makes active visual modules difficult to realize in computer vision. However, the sequential dynamics of decision-making create an opportunity for active visual perception. 
Recognizing the limitations of processing frames in isolation, some recent methods \cite{xu2025vla, li2025sp, wang2025specprune} have begun to leverage historical information, such as frame comparison results and KV-cache reuse, for efficient visual token processing \cite{chen2024image, zhang2024sparsevlm}. However, these approaches mainly focus on visual token pruning for efficiency. Therefore, designing a dynamic, context-aware visual processing paradigm that improves both visual processing and VLA generalization remains a significant challenge.

To address this challenge, we propose AVA-VLA, inspired by the Partially Observable Markov Decision Process (POMDP) framework \cite{smallwood1973optimal, lauri2022partially}. We observe that the core challenge identified above is similar to the POMDP challenge of forming a robust belief state, which functions as a summary of past observations and actions to guide decision-making under uncertainty. 
Since directly computing or representing the belief state is generally intractable, we introduce a recurrent state, which functions as a neural approximation of this belief state and is calculated by the intermediate output of the model in the previous time step. 
%The corresponding graphical model is shown in Figure \ref{fig:figure_1}. 
Then, we design an Active Visual Attention (AVA) module to leverage this recurrent state to calculate the importance of visual tokens and dynamically modulating the visual processing of the current frame. This allows the model to filter and focus its attention based on its historical belief, rather than purely static language instruction. Therefore, the proposed AVA-VLA framework does not rely solely on the current observation but learns to explicitly condition the action prediction on the recurrent state. 

Through extensive experiments in both simulation benchmarks \cite{liu2023libero, fei2025libero, mees2022calvin} and real-world tasks \cite{univla}, we demonstrate that our proposed active visual attention module helps improve policy performance compared to previous VLA frameworks. Our contributions are threefold:

\begin{itemize}
\item We propose the novel AVA-VLA framework to solve the critical limitation of lacking historical context in MDP-based VLA models. To our knowledge, it is the first VLA framework to explicitly address this limitation via a POMDP-inspired approach.
\item We introduce an Active Visual Attention (AVA) module that leverages the recurrent state to dynamically modulate the visual processing of the current frame for action prediction.
\item We conduct comprehensive evaluations in both simulation and real-world tasks, demonstrating that the AVA-VLA framework improves VLA performance, and our method achieves state-of-the-art performance across multiple robot tasks.
\end{itemize}

%% file: sec/2_related_work.tex
\section{Related Work}
\label{sec:formatting}
%-------------------------------------------------------------------------
%\subsection{Vision-Language-Action Modeling}
\noindent\textbf{Vision-Language-Action Models.~}
%\paragraph{Vision-Language-Action Models.}
VLMs \cite{liu2023visual, chen2023pali, karamcheti2024prismatic, shen2024mome} have been pivotal in advancing robotic control by providing rich multi-modal representations. This has fostered the development of VLA models \cite{rt1, rt2, bjorck2025gr00t, cheang2025gr, openvla, pi0, pi0-fast, li2024cogact, song2025reconvla, qian2025geopredict} that bridge high-level perception with low-level action generation. A significant paradigm shift was the introduction of action tokenization by the RT series \cite{rt1, rt2, belkhale2024rt}. This approach treats control as a sequence modeling problem, enabling scalable web-to-robot transfer. Models like OpenVLA \cite{openvla} and UniVLA \cite{univla} generate action policies in the autoregressive (AR) manner. While expressive, the sequential nature of AR decoding is computationally intensive. Therefore, recent research has diversified into more efficient and effective action decoding strategies. Models such as CogACT \cite{li2024cogact} and the $\pi$ series \cite{pi0, pi0-fast}, have explored diffusion-based decoders for the iterative refinement of continuous action trajectories. Other recent works, such as OpenVLA-OFT \cite{openvla-oft} and its variant \cite{li2025cogvla}, employ parallel decoding, which enables the simultaneous prediction of actions within the action chunk, improving inference efficiency and supporting scalable deployment. %Concurrently, the field has been advanced by large-scale data aggregation efforts, such as the Octo \cite{octo} dataset, which supports robust multi-task training. 

%-------------------------------------------------------------------------
\noindent\textbf{Sequential Processing in VLMs.~}
%\paragraph{History-aware Vision-Language Models.}
Many VLM studies \cite{qian2024streaming, pan2025semantic, wang2025continuous, fan2025vlm} focus on processing sequential visual data for tasks such as video understanding and temporal-based video questions. These works efficiently aggregate historical information, allowing the model to build a holistic, temporal-aware representation of the video's content. VLM-3R \cite{fan2025vlm} employs a geometry encoder to derive implicit 3D tokens that represent spatial-temporal understanding. \cite{wang2025continuous} incrementally updates a persistent internal state that encodes the scene content.  These VLM models use history for passive comprehension or offline understanding. In contrast, VLA models operate in an active, dynamic decision-making environment, which requires the model to interact with the environment. This distinction motivates our POMDP-inspired approach, which focuses on maintaining a recurrent state for active decision-making. 
%, rather than simply aggregating temporal features for passive comprehension.

%-------------------------------------------------------------------------
%\textcolor{blue}{\subsection{diffusion action policy}}

%% file: sec/3_method.tex
\section{Methods}
\begin{figure*}[tbp]
\centering
\includegraphics[width=0.95\linewidth]{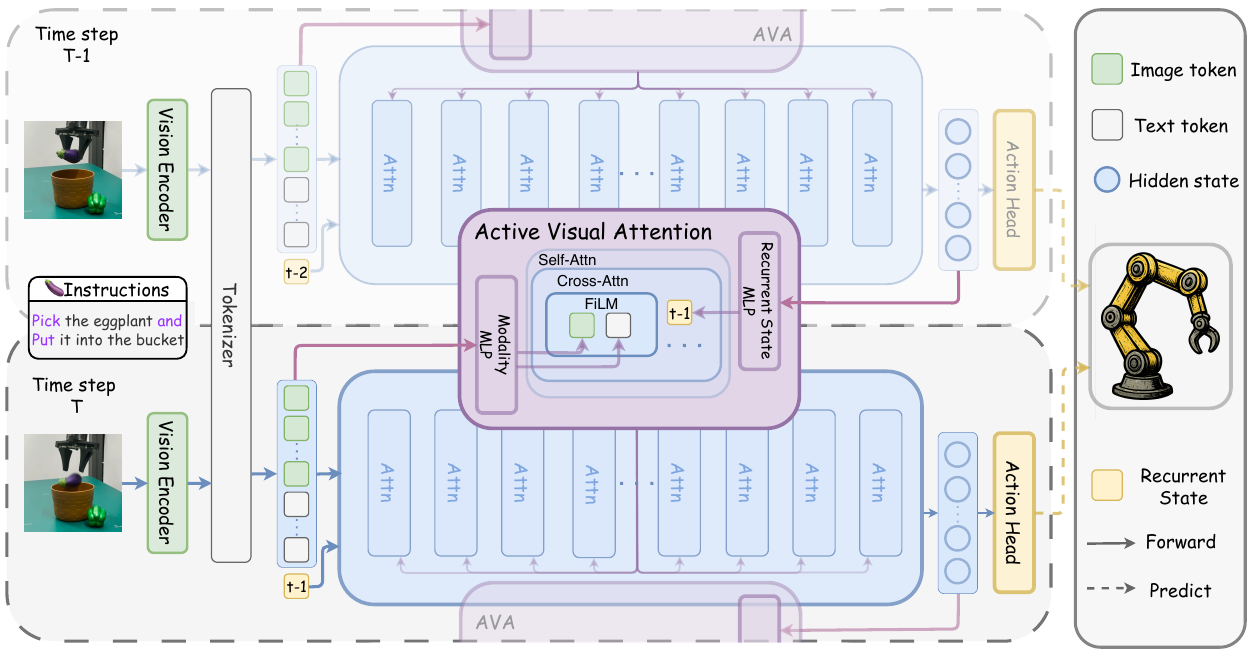}
\caption{Overview of the proposed AVA-VLA framework. At each timestep, the recurrent state is projected from the previous hidden state to preserve historical context and to initialize the current action tokens. Then the AVA module combines this recurrent state with text-conditioned visual features from the current observation to generate soft importance scores, which modulate the visual attention matrices throughout the backbone LLM, enabling the model to focus on task-relevant regions based on both temporal context and current perception.}
% For two adjacent observations in a temporal sequence, the hidden state of the former observation from the LLM backbone is projected into an action embedding placeholder of the latter. The resulting cross-attention between text-conditioned visual features and the recurrent state serves as a soft importance score to modulate the visual attention matrices at each backbone layer. Together with the AVA module, the recurrent state enhances temporal modeling across observations.
\label{fig:framework}
\vskip -0.1in
\end{figure*}

In this section, we present our proposed VLA method. We begin with the preliminaries (\ref{sec:method_pre}), followed by the AVA-VLA framework (\ref{sec:method_framework}) and the detailed description of the proposed Active Visual Attention module (\ref{sec:method_AVA}). We then explain our training and inference procedures (\ref{sec:method_train}). An overview of the proposed AVA-VLA framework is shown in Figure \ref{fig:framework}.

\subsection{Preliminaries}\label{sec:method_pre}
A typical VLA model $\mathcal{P}_{\theta}$, parameterized by $\theta$, consists of four main components: a Large-Language-Model (LLM) backbone $\mathcal{M}$, a vision encoder $\mathcal{E}$, a language tokenizer $\mathcal{T}$, and an action head (or de-tokenizer) $\mathcal{Q}$. We thus define the model as $\mathcal{P}=\{\mathcal{M}, \mathcal{E}, \mathcal{T}, \mathcal{Q}\}$.

Following the representative OpenVLA \cite{openvla}, at timestep $t$, given an input tuple $\boldsymbol{x}^t = (\boldsymbol{x}_I^t, \boldsymbol{x}_S^t)$, the visual encoder $\mathcal{E}$ first encodes the input image $\boldsymbol{x}_I^t$ into $\mathrm{L}_I$ visual tokens: $\boldsymbol{z}_I^t = \mathcal{E}(\boldsymbol{x}_I^t) \in \mathbb{R}^{\mathrm{L}_I \times d}$, where $d$ denotes the embedding dimension.
These visual tokens are then concatenated with $\mathrm{L}_S^t$ language tokens, $\boldsymbol{z}_S^t = \mathcal{T}(\boldsymbol{x}_S^t) \in \mathbb{R}^{\mathrm{L}_S^t \times d}$.
The combined sequence is then fed into the LLM backbone $\mathcal{M}$ to generate output hidden states $\boldsymbol{h}^t$.
Finally, the action head $\mathcal{Q}$ maps the output hidden states $\boldsymbol{h}^t$ into a $D$-dimensional executable action $\mathcal{A}^t$ for robotic control (e.g., $D=7$ for 3-DoF translation, 3-DoF rotation, and binary gripper control).
Thus, the entire forward pass at timestep $t$ can be formulated as:
\begin{align}
    \mathcal{A}^t = \mathcal{Q}(\boldsymbol{h}^t) = \mathcal{Q}(\mathcal{M}(\boldsymbol{z}_I^t, \boldsymbol{z}_S^t)).
\end{align}

% To improve the computation overhead of autoregressive (AR) decoding in VLA's sequenced action prediction, OpenVLA-OFT \cite{openvla-oft} and its variant \cite{li2025cogvla} employ parallel decoding, which enables the simultaneous prediction of all actions within the action chunk, improving inference efficiency and supporting scalable deployment. 

Recent representative VLA models, such as OpenVLA-OFT \cite{openvla-oft} and its variant \cite{li2025cogvla}, map the output hidden states into an executable action chunk $\mathcal{A}^t = [a_0^t, a_1^t, ..., a_{\mathrm{L}_c-1}^t] \in \mathbb{R}^{\mathrm{L}_c \times D}$, where $\mathrm{L}_c$ and $D$ represent the length of the action chunk and the dimensionality of each atomic action, respectively. 
To facilitate parallel generation, a learnable action placeholder embedding $\boldsymbol{p}^t$ is appended to the input sequence \cite{li2025cogvla, lin2024petformer}. This placeholder embedding is set to empty in OpenVLA-OFT, i.e., $\boldsymbol{p}^t = \bar{\boldsymbol{0}}=[\boldsymbol{0}_0,\boldsymbol{0}_1,...,\boldsymbol{0}_{\mathrm{L}_c-1}] \in \mathbb{R}^{\mathrm{L}_c \times D \times d}$.
% \begin{align}
%     \hat{\boldsymbol{z}}_I^t = \mathcal{E}_{\text{FiLM}}(\boldsymbol{x}_I^t, \boldsymbol{z}_S^t ),
% \end{align}
% where $\mathcal{E}_{\text{FiLM}}$ is the vision encoder augmented with the FiLM operator.
The corresponding forward pass under parallel decoding at timestep $t$ can thus be expressed as:
\begin{align}
    \mathcal{A}^t = \mathcal{Q}(\mathcal{M}_{\text{parallel}}(\boldsymbol{z}_I^t, \boldsymbol{z}_S^t, \boldsymbol{p}^t)).
\end{align}
Regardless of whether AR or parallel decoding is used, these VLA models learn to predict the action $\bar{\mathcal{A}}^t$ only from the current observation $\boldsymbol{x}^t$. This implicitly models the task as a Markov decision process:
\begin{align}\label{eq:mdp}
    \bar{\mathcal{A}}^t \sim \mathcal{P}_{\theta}(\mathcal{A}^t \mid \boldsymbol{x}^t).
\end{align}

\subsection{AVA-VLA Framework}\label{sec:method_framework}

The history-agnostic design for policy learning in Eq.~\eqref{eq:mdp} is suboptimal for effective visual token processing in dynamic sequential decision-making \cite{zhang2017dynamically, lauri2022partially}, as it fails to capture non-observable dynamics or occluded information. 
This limitation inspired us to re-formulate the VLA model from a POMDP perspective. In a POMDP framework, the optimal policy at timestep $t$ should be conditioned not only on the current observation $\boldsymbol{x}^t$ but also on a belief state $b^{t-1}$, which captures all relevant historical context, including observations and actions, i.e., $b^{t-1} = P(s_{t-1} \mid \boldsymbol{x}^{<t}, \mathcal{A}^{<t})$. Inspired by this theoretical framework, we re-formulate the VLA policy as
\begin{align}
    \bar{\mathcal{A}}^t \sim \mathcal{P}_{\theta}(\mathcal{A}^t \mid \boldsymbol{x}^t, b^{t-1}).
\end{align}

This formulation provides a theoretical foundation for designing a more effective visual processing paradigm, suggesting that leveraging historical context in observations can improve VLA generalization.
Since computing the theoretical belief state $b^{t-1}$ is generally intractable, we instead propose to learn a compressed representation, $\boldsymbol{r}^{t-1}$, as its neural approximation. 
This approach naturally transforms the VLA model into a recurrent structure \cite{medsker2001recurrent}, leading to a non-Markovian policy conditioned on this learned representation: $\bar{\mathcal{A}}^t \sim \mathcal{P}_{\theta}(\mathcal{A}^t \mid \boldsymbol{x}^t, \boldsymbol{r}^{t-1})$.

In our proposed AVA-VLA framework, we term this approximate vector $\boldsymbol{r}^{t-1}$ the recurrent state, which captures historical context. In typical VLA models, the hidden states immediately preceding action generation contain fused visual and language information and are predictive of the agent's intent. Therefore, we derive the recurrent state for timestep $t$ from the action-related hidden state at timestep $t-1$.

Specifically, for a parallel-decoding-based VLA model, which contains $M$ decoder layers that predict $\mathrm{L}_{A}=\mathrm{L}_cD$ actions in one forward pass, we denote its hidden states output at the $m$-th layer and time $t$ by $h_m^{t} \in \mathbb{R}^{\mathrm{L}_A \times d}$. The corresponding recurrent state is computed by:
\begin{align}
    \boldsymbol{r}^{t-1} = \mathcal{B}(\boldsymbol{h}_{M}^{t-1}) \in \mathbb{R}^{\mathrm{L}_A \times d},
\end{align}
where $\mathcal{B}$ is an MLP module that transforms the hidden state into the recurrent state. 

We employ this recurrent state to guide the VLA model to actively focus on visual regions that become critical along the time sequence.
To utilize the recurrent state, we introduce the active visual attention module by quantifying the importance of visual tokens and dynamically modulating the processing of the visual frame for the current timestep. Moreover, in order to preserve the rich historical information, we use this recurrent state $\boldsymbol{r}^{t-1}$ for action placeholder \cite{openvla-oft, li2025cogvla} embedding initialization, i.e., $\boldsymbol{p}^t = \boldsymbol{r}^{t-1}$.

For simplicity, our framework is built upon the OpenVLA-OFT foundation model. 
Therefore, the forward pass at timestep $t$, incorporating the AVA module and state-based initialization, is formulated as:
\begin{align}\label{eq:ava_vla}
    \mathcal{A}^t = \mathcal{Q}(\mathcal{M}_{\text{parallel}}(\boldsymbol{z}_I^t, \mathcal{V}(\boldsymbol{x}^t, \boldsymbol{r}^{t-1}), \boldsymbol{z}_S^t, \boldsymbol{r}^{t-1})),
\end{align}
where $\mathcal{V}$ is the proposed AVA module, which takes the current observations and the recurrent state as input.

\subsection{Active Visual Attention}\label{sec:method_AVA}
We now describe the detailed architecture of the AVA module $\mathcal{V}$, which is designed to modulate visual processing in a dynamic manner.

% The AVA module first employ feature-wise linear modulation (FiLM) \cite{perez2018film} to condition the visual features on the language instruction, i.e., $\bar{\boldsymbol{z}}_I^t = \mathcal{F}_{\gamma}(\boldsymbol{z}_S^t) \odot \boldsymbol{z}_I^t + \mathcal{F}_{\beta}(\boldsymbol{z}_S^t)$. Then uses the vision tokens $\hat{\boldsymbol{z}}_I^t$ as the query, 
% \begin{align}
%     \mathbf{Q}^t = W_Q\hat{\boldsymbol{z}}_I^t \in \mathbb{R}^{\mathrm{L}_I\times d'},
% \end{align}
% and the recurrent state $\boldsymbol{r}^{t-1}$ as the key and value
% \begin{align}
%     \mathbf{K}^t, \mathbf{V}^t = (W_K/W_V)\boldsymbol{r}^{t-1} \in \mathbb{R}^{\mathrm{L}_A\times d'},
% \end{align}
The AVA module first employs modality-specific MLPs to encode the visual features $\boldsymbol{z}_I^t$ and the instruction feature $\boldsymbol{z}_S^t$ into $\bar{\boldsymbol{z}}_I^t\in \mathbb{R}^{\mathrm{L}_I\times d'}$ and $\bar{\boldsymbol{z}}_S^t\in \mathbb{R}^{\mathrm{L}_S^t\times d'}$, respectively, where $d' <d$. A feature-wise linear modulation (FiLM) \cite{perez2018film} is applied to condition the visual features on the language instruction, i.e., $\hat{\boldsymbol{z}}_I^t = \mathcal{F}_{\gamma}(\bar{\boldsymbol{z}}_S^t) \odot \bar{\boldsymbol{z}}_I^t + \mathcal{F}_{\beta}(\bar{\boldsymbol{z}}_S^t)$. Then it uses the visual tokens $\hat{\boldsymbol{z}}_I^t$ as the query, 
\begin{align}
    \mathbf{Q}^t = W_Q\hat{\boldsymbol{z}}_I^t \in \mathbb{R}^{\mathrm{L}_I\times d'},
\end{align}
and the recurrent state as the key and value
\begin{align}
    \mathbf{K}^t, \mathbf{V}^t = (W_K/W_V)\hat{\boldsymbol{r}}^{t-1} \in \mathbb{R}^{\mathrm{L}_A\times d'},
\end{align}
where $W_Q$, $W_K$, $W_V$ are linear projection layers, and $\hat{\boldsymbol{r}}^{t-1} \in \mathbb{R}^{\mathrm{L}_A\times d'}$ is the output of $\boldsymbol{r}^{t-1}$ after MLP encoding. Then it computes the attention matrix and feeds the output into a self-attention layer
\begin{align}
    \mathbf{O}^t = \text{Self-Att}\left(\text{Cross-Att}(\mathbf{Q}^t, \mathbf{K}^t,\mathbf{V}^t)\right).
\end{align}
Inspired by \cite{rao2021dynamicvit, sun2025lvpruning}, we feed the resulting tokens $\mathbf{O}^t\in \mathbb{R}^{\mathrm{L}_I \times d'}$ into a Feedforward Neural Network (FFN), a linear layer $\mathcal{W}: \mathbb{R}^{d'} \to \mathbb{R}^2$, and apply a Softmax function along the feature dimension. This predicts the logits for enhancing or weakening each visual token as:
\begin{align}
    \boldsymbol{\rho}^t = \text{Softmax}\left( \mathcal{W}\left(\text{FFN} \left( \mathbf{O}^t\right)\right)\right)  \in \mathbb{R}^{\mathrm{L}_I \times 2}.
\end{align}
Then we compute the final soft weights for visual tokens at time $t$ by $\boldsymbol{\omega}^t = \boldsymbol{\rho}^t\boldsymbol{\gamma}$, where $\boldsymbol{\gamma}$ is a 2-dimensional vector. 
These soft weights directly represent the importance scores of the visual tokens. The components of $\boldsymbol{\gamma}$, $\gamma_0$ and $\gamma_1$, represent the scalar scores for enhancing and weakening a visual token, respectively.

The soft weights vector $\boldsymbol{\omega}^t$ is applied to all layers of the LLM backbone. Specifically, at time step $t$, let the total sequence length be $\textrm{L}_{o}^t$, we denote the attention score of the $m$-th layer by $\mathbf{C}^{t,m}\in \mathbb{R}^{\textrm{L}_o^t \times \textrm{L}_o^t}$, which is calculated by applying the original attention mask to the raw attention scores.
The final attention matrix of the $m$-th layer $\mathbf{A}^{t,m}$ is calculated by applying the soft attention mask matrix $\mathbf{U}^t$ with the $\textit{Softmax}$ operation to $\mathbf{C}^{t,m}$:
\begin{align}
    \mathbf{A}_{i,j}^{t,m}=\frac{\exp(\mathbf{C}_{i,j}^{t,m})\mathbf{U}_{i,j}^t}{\sum_{l=1}^{\textrm{L}_{o}^t}\exp(\mathbf{C}_{i,l}^{t,m})\mathbf{U}_{i,l}^t}, ~1\le i,j \le \textrm{L}_{o}^t,
\end{align}
where the soft attention matrix $\mathbf{U}^t$ is constructed based on the soft weights vector $\boldsymbol{\omega}^t$:
\begin{align}
\mathbf{U}^t_{i,j} = 
    \begin{cases} 
    1 & \text{if } i = j \text{ or } j \not\in \Lambda_{I}, ~~~~ 1 \leq i, j \leq \textrm{L}_{o}^t\\
    \boldsymbol{\omega}^t_j & \text{if } i \neq j  \text{ and } j \in \Lambda_{I}, ~1 \leq i, j \leq \textrm{L}_{o}^t
    \end{cases},
\end{align}
where the set $\Lambda_{I}$ represents the indices of the visual tokens.

Therefore, the proposed AVA module uses the recurrent state and current visual observation to calculate soft weights to guide the VLA model to filter and focus its attention based on historical information.

% \begin{figure}[!htbp]
% \centering
% \includegraphics[width=1\linewidth]{fig/Test_2.png}
% \caption{..
% }
% \label{fig:mcs}
% \end{figure}

\subsection{Training and Inference Procedure}\label{sec:method_train}
The proposed AVA-VLA framework introduces a recurrent dependency through the recurrent state $\boldsymbol{r}^{t-1}$. Training such a recurrent model ideally requires backpropagation through time over the entire trajectory to capture long-term dependencies. However, given the substantial memory constraint and computational cost of modern VLA backbones, performing the full backpropagation through time is computationally prohibitive \cite{pascanu2013difficulty}.

To address this challenge, we adopt a truncated backpropagation through time strategy \cite{liao2018reviving}. We unroll the model for a fixed, short horizon. Specifically, for the $n$-th sample in the training batch, it contains a continuous observation sequence $\{\boldsymbol{x}^{t,n}\}_{t=0}^{T-1}$. For each timestep $t$ in this sequence, we calculate the action chunk prediction loss using the Mean Absolute Error (MAE): $\mathcal{L}^{t,n} = \mathcal{L}(\mathcal{A}^{t,n}, \mathcal{A}_{\text{GT}}^{t,n})$. To prevent overly dispersed soft attention weights, we add an L2 penalty regularizer $\mathcal{L}_{\omega}^{t,n}$ on the mean value of the weight vector $\boldsymbol{\omega}^{t,n}$, defined as:
\begin{align}
    \mathcal{L}_{\omega}^{t,n} = \|\mu(\boldsymbol{\omega}^{t,n}) - c\|,
\end{align}
where $\mu(\cdot)$ is the mean function and $c$ is a target mean hyperparameter. This encourages the AVA module to focus on task-relevant regions while suppressing distracting background responses (\textit{see Appendix \ref{sec:app_add} for more analysis}).

Therefore, the total loss of one training batch is the sum of the prediction loss and penalty loss  of $N$ truncated sequences:
\begin{align}
    \mathcal{L}_{\text{total}} = \sum\nolimits_{n=1}^{N} \sum\nolimits_{t=0}^{T-1} (\mathcal{L}^{t,n} + \lambda \mathcal{L}_{\omega}^{t,n}),
\end{align}
where $N$ is the batch size, and $\lambda$ is a balancing coefficient. In our experiments, we set $T=4$ to balance computational feasibility with the need to learn the temporal dynamics captured by the recurrent state. At the first timestep ($t=0$) of any sequence, the initial recurrent state $\boldsymbol{r}^{-1}$ is initialized as a zero embedding, i.e., $\boldsymbol{r}^{-1} = \bar{\boldsymbol{0}}$. 

During inference, the model operates in a fully recurrent manner. At the beginning of a new episode ($t=0$), the initial recurrent state $\boldsymbol{r}^{-1}$ is initialized as the zero embedding. Then, for each subsequent timestep $t \geq 0$, the agent receives the current observation $\boldsymbol{x}^t$ and performs a single forward pass as defined in Eq.~\eqref{eq:ava_vla}, conditioned on both $\boldsymbol{x}^t$ and the previously computed recurrent state $\boldsymbol{r}^{t-1}$. This forward pass predicts the action chunk $\mathcal{A}^t$ and simultaneously extracts the recurrent state $\boldsymbol{r}^{t}$. This loop continues for the entire inference process.

\noindent\textbf{Remark.~}
%\paragraph{Remark.} 
We explicitly note that the soft weights vector $\boldsymbol{\omega}^t$ computed by the AVA module has a natural application in visual token reduction \cite{wang2023efficientvlm, li2025sp}. 
Visual tokens with low importance scores can be pruned to reduce the computational cost of the LLM backbone. 
While this is a valid direction for improving model efficiency, it is not the primary focus of this work. We provide a preliminary exploratory analysis on leveraging the weight vector to do token reduction, which further validates the effectiveness of our proposed method. Details can be found in Section \ref{sec:analysis}.

%% file: sec/4_experiment.tex
\begin{table*}[!th]
% \small
\caption{Comparison on the LIBERO benchmark. The results are reported in two groups: one policy for all 4 suites, and one policy per suite. The best results in each column of each group are highlighted in \textbf{bold}.
}
\vskip -0.2in
\label{tab:sota_exps}
\begin{center}
\resizebox{0.8\textwidth}{!}{
\begin{tabular}{l|cccc|c}
\toprule
Method & Spatial SR (\%) & Object SR (\%) & Goal SR (\%) & Long SR (\%) & Average SR (\%)\\ \midrule
 \multicolumn{6}{c}{\textit{One policy for all 4 suites}} \\
  TraceVLA \cite{tracevla} & 84.6 & 85.2 & 75.1 & 54.1 & 74.8 \\
  WorldVLA \cite{cen2025worldvla}& 87.6&	96.2	&83.4	&60.0&	81.8  \\
$\pi_0$ \cite{pi0} & 96.8 & 98.8 & 95.8 & 85.2 & 94.2 \\
$\pi_0$-FAST \cite{pi0-fast} & 96.4 & 96.8 & 88.6 & 60.2 & 85.5 \\
UnifiedVLA \cite{wang2025unifiedvisionlanguageactionmodel}& 95.4&	98.8	&93.6	&94.0	&95.5\\ 
OpenVLA-OFT \cite{openvla-oft} & \textbf{97.7} &98.0 &96.1 &95.3 &96.8\\ 
    % \textcolor{gray}{GeoVLA \cite{geovla}\textcolor{gray}{\small \textit{[arXiv'25]}}} & \textcolor{gray}{98.4} & \textcolor{gray}{99.0} & \textcolor{gray}{96.6} & \textcolor{gray}{96.6} & \textcolor{gray}{97.7} \\
    % \textcolor{gray}{3D-CAVLA \cite{bhat20253d}\textcolor{gray}{\small \textit{[arXiv'25]}}} & \textcolor{gray}{98.2} & \textcolor{gray}{99.8} & \textcolor{gray}{98.2} & \textcolor{gray}{96.1} & \textcolor{gray}{98.1} \\
\midrule
        AVA-VLA (Ours)  & 97.4 & \textbf{99.4} & \textbf{97.4} & \textbf{97.6} & \textbf{98.0} \\
      \midrule [1.2pt]
       \multicolumn{6}{c}{\textit{One policy per suite}} \\
% Diffusion Policy \cite{diffusion-policy} & 78.3 & 92.5 & 68.3 & 50.5 & 72.4 \\
% Octo \cite{octo} & 78.9 & 85.7 & 84.6 & 51.1 & 75.1 \\
OpenVLA \cite{openvla} & 84.7 & 88.4 & 79.2 & 53.7 & 76.5 \\
SpatialVLA \cite{qu2025spatialvla} & 88.2 & 89.9 & 78.6 & 55.5 & 78.1 \\
% Dita \cite{dita}& 84.2 & 96.3 & 85.4 & 63.8 & 82.4 \\
CoT-VLA \cite{cot-vla} & 87.5 & 91.6 & 87.6 & 69.0 & 83.9 \\
NORA \cite{hung2025nora} &92.2	&95.4&	89.4	&74.6&	87.9  \\
PD-VLA \cite{song2025accelerating}&95.5&	96.7	&94.9&	91.7	&94.7 \\
UniVLA \cite{univla} & 96.5 & 96.8 & 95.6 & 92.0 & 95.2 \\
OpenVLA-OFT \cite{openvla-oft} & 97.6 & 98.4 & 97.9 & 94.5 & 97.1 \\ 
FLOWER \cite{reuss2025flower}&97.5& 99.1 &96.1 & 94.9  & 96.9 \\
% VLA-Adapter \cite{wang2025vla} & 97.8 & 	99.2 & 	97.2 & 	95.0	& 97.3 \\
RIPT-VLA \cite{tan2025interactive}& 99.0 &	98.6 &	\textbf{98.6} &	93.8 &	97.5  \\
  \midrule
    AVA-VLA (Ours)& \textbf{99.2} & \textbf{99.6} & 97.9 & \textbf{96.2} & \textbf{98.2} \\
  \bottomrule
\end{tabular}
}
\end{center}
\vskip -0.2in
\end{table*}

\section{Experiments}
We evaluate the effectiveness of our approach through a set of experiments spanning both simulation benchmarks and real-world robot manipulation tasks. Additionally, we conduct a comprehensive ablation study and analysis to validate the effectiveness of our approach. All experiments are conducted on Nvidia A800 GPUs. 
\subsection{Experimental Setup}

We conduct experiments on three challenging settings: the LIBERO \cite{liu2023libero} and CALVIN \cite{mees2022calvin} benchmarks for evaluation in simulation environments, and a real-world table-mounted Mobile ALOHA robot with four test tasks, to validate the sim-to-real transferability of our method. We use the open-source OpenVLA-OFT \cite{openvla-oft} as our foundation model, which consists of a two-branch vision encoder (DINOv2 and SigLIP) and a LLaMA2-7B backbone \cite{touvron2023llama}. \textit{Due to space limitations, implementation details are provided in Appendix \ref{sec:app_imp}.}

\noindent\textbf{LIBERO.} % Benchmark.~
%\noindent\textbf{LIBERO.~}
%\paragraph{LIBERO.}
% LIBERO \cite{liu2023libero} is a benchmark for lifelong robot learning, designed to study knowledge transfer in multitask settings. It uses a Franka Emika Panda arm in MuJoCo simulations, with datasets split into four suites: LIBERO-Spatial (varying layouts), LIBERO-Object (diverse objects), LIBERO-Goal (goal-conditioned), and LIBERO-Long (long-horizon). Each suite has 10 tasks with 50 demonstrations, totaling ~5,000 episodes across 100 tasks. Data includes RGB images (200x200 resolution), proprioceptive states, and delta actions. 
% It supports offline imitation and online reinforcement, with procedural generation for diversity. 
LIBERO \cite{liu2023libero} is a benchmark for lifelong robot learning. It uses a Franka Emika Panda arm in MuJoCo, with datasets split into four suites: LIBERO-Spatial, LIBERO-Object, LIBERO-Goal, and LIBERO-Long. It contains 5,000 episodes across 100 tasks. Data includes RGB images, proprioceptive states, and delta actions, with procedural generation for diversity.
\textit{LIBERO+ \cite{fei2025libero} is a challenging LIBERO-based benchmark, which offers a robust benchmarking framework with 7 perturbation dimensions and 21 sub-dimensions. It allows users to assess model performance across various challenges systematically. We conduct additional experiments on the LIBERO+ benchmark, and the results are put in Appendix \ref{sec:app_add}.}
% 精简版本备选：
% \noindent\textbf{LIBERO.}
% LIBERO \cite{liu2023libero} is a benchmark for lifelong robot learning. It uses a Franka Emika Panda arm in MuJoCo, with datasets split into four suites: LIBERO-Spatial (varying layouts), LIBERO-Object (diverse objects), LIBERO-Goal (goal-conditioned), and LIBERO-Long (long-horizon). It contains 5,000 episodes across 100 tasks. Data includes RGB images, proprioceptive states, and delta actions, with procedural generation for diversity.
% LIBERO+ \cite{fei2025libero} extends LIBERO with systematic perturbations for robustness evaluation. We conduct additional experiments on LIBERO+, with results in Appendix \ref{sec:app_add}.

\noindent\textbf{CALVIN.} % Benchmark.~
%\paragraph{CALVIN.} 
% CALVIN \cite{mees2022calvin} is a simulated benchmark for language-conditioned, long-horizon manipulation, using a Franka Panda arm with RGBD observations (third-person 200x200, wrist 80x80), proprioception, and natural language goals. It evaluates sequential reasoning in VLA. CALVIN spans 34 tasks across four environments (A-D), with 20,000+ episodes from teleoperated play data, emphasizing unseen object generalization and multi-stage sequences (e.g., ``open drawer, pick blue block, push into drawer"). Following \cite{univla}, we used the CALVIN ``ABC$\to$D" setting, which means training on environments A, B, and C and evaluating on environment D, to evaluate performance on the zero-shot generalization tasks. 
CALVIN \cite{mees2022calvin} is a simulated benchmark for language-conditioned, long-horizon manipulation, using a Franka Panda arm with RGBD observations, proprioception, and natural language goals. It evaluates sequential reasoning in VLA. CALVIN spans 34 tasks across four environments (A-D), with 20,000+ episodes, emphasizing unseen object generalization and multi-stage sequences (e.g., ``open drawer, pick blue block, push into drawer"). Following \cite{univla}, we used the CALVIN ``ABC$\to$D" setting, which means training on environments A, B, and C and evaluating on environment D, to evaluate performance on the zero-shot generalization tasks. 

\noindent\textbf{Mobile ALOHA Real-Robot Experiments.~} % Real-Robot Experiments.~
%\paragraph{Mobile ALOHA Real-Robot Experiments.} 
We use a stationary cobot magic dual-arm robot to assess our model’s adaptability to novel real-world environments with a small number of robot demonstrations. Following \cite{univla, cot-vla}, we perform evaluations across four challenging tasks. These include Pick and Place, which involves placing the bucket in the center, and then placing irregular-shaped objects into the bucket (e.g., ``put $<$obj$>$ into bucket"), and Sequenced Instruction Understanding, which requires executing multi-step commands like stacking a Tower of Hanoi (``Stack tower of hanoi”). We also test Flexible Object Folding, a deformable object manipulation task requiring a specific three-stage process to fold a towel (``fold towel twice"), and Dexterous Action, which involves fine-motor skills such as using a shovel to scoop small items (e.g., corn, sesame seeds) into a bowl. For each task, the dataset contains between 30 and 450 demonstrations. \textit{Details of the task suites are provided in Appendix \ref{sec:app_exp}.}

\subsection{Evaluation Results}
\begin{table}[!tbp]
    \centering
	\caption{
    Comparison on the CALVIN ABC$\to$D benchmark. The results are reported in terms of success rates (\%) and average length. The best results in each column are highlighted in \textbf{bold}.}\label{ComparisonCALVIN}%
    \vskip -0.1in
    \resizebox{0.48\textwidth}{!}{
\begin{tabular}{l|ccccc|c}
\toprule
    CALVIN   & \multicolumn{5}{c|}{~ Task completed in a row $\uparrow$} & Avg. len \\
     ABC$\to$D & 1 & 2 & 3 & 4 & 5 &$\uparrow$  \\
\midrule
OpenVLA \cite{openvla} & 91.3  & 77.8  & 62.0    & 52.1  & 43.5  & 3.27 \\
UniVLA \cite{univla} & 95.5  & 85.8  & 75.4    & 66.9  & 56.5  & 3.80 \\
UnifiedVLA \cite{wang2025unifiedvisionlanguageactionmodel}& 98.9 & 94.8 & 89.0 & 82.8 & 75.1 & 4.41 \\
OpenVLA-OFT \cite{openvla-oft} &96.9&92.0&85.7&80.4&72.9&4.28\\
FLOWER \cite{reuss2025flower}& 99.4 & 95.8 & 90.7 & 84.9 & 77.8 & 4.53 \\
VLA-Adapter \cite{wang2025vla}& 	99.1& 94.6& 88.8& 82.8& 76.5& 4.42 \\
Seer \cite{tian2024predictive} & 96.3&	91.6	&86.1&	80.3&	74.0	&4.28 \\
    \midrule
AVA-VLA (Ours) & \textbf{99.6} & \textbf{97.6} & \textbf{94.1} & \textbf{89.9} & \textbf{84.1} & \textbf{4.65} \\
  \bottomrule
\end{tabular}}
\vskip -0.2in
\end{table}

\begin{figure*}[!t]
\centering
\includegraphics[width=0.95\linewidth]{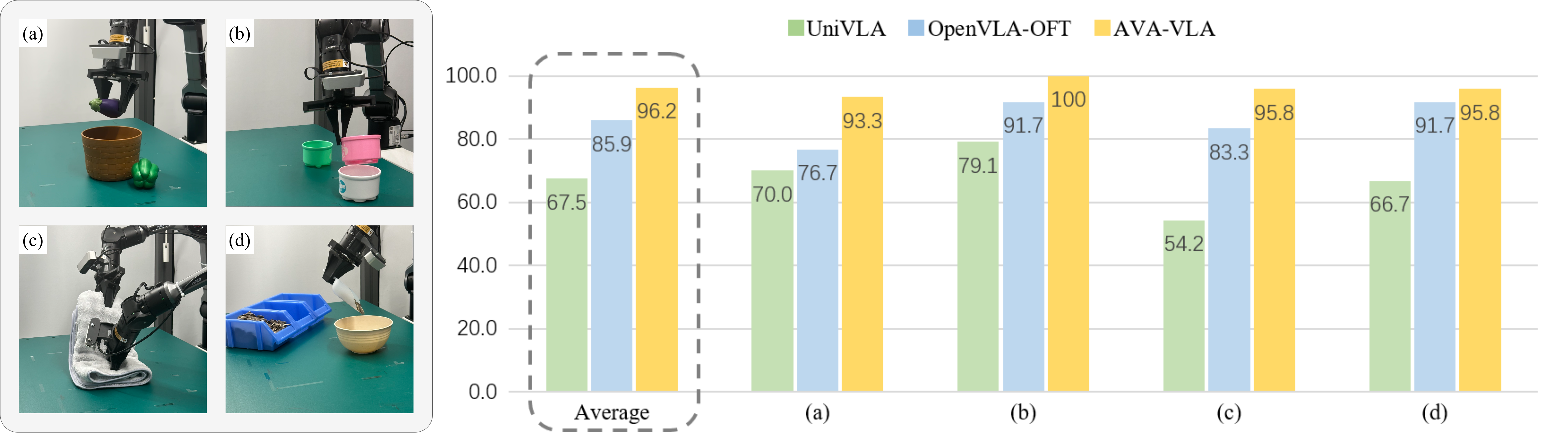}
\vskip -0.01in
\caption{Comparison on the Mobile ALOHA real-world experiments. Evaluation across four manipulation tasks, including (a) Pick and Place, (b) Sequenced Instruction Understanding, (c) Flexible Object Folding, (d) Dexterous Action. Left: Representative middle states for each task setup. Right: Task-specific success rates and cross-task averages for our method and baselines.
}
\label{fig:figure_real_main}
\vskip -0.2in
\end{figure*}

\noindent\textbf{Baselines.~}
%\paragraph{Baselines.} 
We selected recently published works' main method as baselines. They are TraceVLA \cite{tracevla}, WorldVLA \cite{cen2025worldvla}, $\pi_0$ \cite{pi0}, $\pi_0$-FAST \cite{pi0-fast}, UnifiedVLA \cite{wang2025unifiedvisionlanguageactionmodel}, OpenVLA-OFT \cite{openvla-oft}, OpenVLA \cite{openvla}, SpatialVLA \cite{qu2025spatialvla}, CoT-VLA \cite{cot-vla}, NORA \cite{hung2025nora}, PD-VLA \cite{song2025accelerating}, UniVLA \cite{univla}, OpenVLA-OFT \cite{openvla-oft}, FLOWER \cite{reuss2025flower}, RIPT-VLA \cite{tan2025interactive}, VLA-Adapter \cite{wang2025vla}, Seer \cite{tian2024predictive}. The results of these baselines in LIBERO and  CALVIN benchmarks are based on original references or other published works, ensuring objectivity and correctness. For Mobile ALOHA real-robot experiments, we select UniVLA and OpenVLA-OFT methods as baselines.

\noindent\textbf{Evaluation Metrics.~}
%\paragraph{Evaluation Metrics} 
We use widely adopted performance evaluation metrics ``Success Rate (SR)” (the same in LIBERO \cite{liu2023libero}) to evaluate the results for three challenging settings. In addition, we use ``Average len” of completed tasks (the larger the better, with values between 0-5) as metrics for the CALVIN benchmark.

% \noindent\textbf{Implementation Details.~}
% %\paragraph{Implementation Details.}
% All experiments are conducted on Nvidia A100 GPUs. For the proposed AVA-VLA framework, we use the open-sourced OpenVLA-OFT as our foundation model, which consists of a two-branch vision encoder (DINOv2 and SigLIP) and an LLaMA-2-7B backbone. Following \cite{openvla-oft}, the model is initialized from the official OpenVLA checkpoints. To fine-tune the model, we apply the LoRA technique \cite{hu2022lora} with a rank of 32 to the vision encoder and LLM backbone, while the action head and proprioceptive projector, and the proposed AVA module are fully fine-tuned. The length of the observation sequence is set to $K=4$. The model is trained for 150,000 gradient steps with an initial learning rate of 5e-4, which includes a warm-up phase from 10\% of the value for stability. The learning rate is decayed to 5e-5 after 100,000 steps. \textit{The implementation details of Mobile ALOHA real-robot experiments are put in Appendix \ref{sec:app_imp}.}

\noindent\textbf{LIBERO.~}
%\paragraph{LIBERO.} 
We present quantitative results in Table \ref{tab:sota_exps}. Following established baselines, we conduct experiments in two different settings: training policies on each task suite independently (single-task learning) and training a single policy for all task suites (multi-task learning \cite{linreasonable}). Results demonstrate that the proposed AVA-VLA framework achieves state-of-the-art overall performance in both single-task and multi-task settings. Moreover, it consistently achieves the best performance on the most challenging LIBERO-Long task suite. These results demonstrate the superiority of the proposed AVA-VLA framework.

\noindent\textbf{CALVIN.~}
%\paragraph{CALVIN.}
We present the success rates for each task and the average completed length across all five tasks of the CALVIN benchmark in Table \ref{ComparisonCALVIN}. The results show that the proposed AVA-VLA framework comprehensively outperforms baseline methods across all tasks. This demonstrates our method's strong generalization ability, with an average length superior to previous state-of-the-art baselines. 

%From the perspective of training module, the advantage stems from our implicit grounding learning framework, which enables the model to precisely attend to visual information at target objects, thereby achieving precise manipulation.

\noindent\textbf{Mobile ALOHA.~}
%\paragraph{Mobile ALOHA.} 
The Pick and Place task is evaluated for a total of 30 trials (10 per object), while other tasks are evaluated for 24 trials each. The experimental results on real-world tasks are reported in Figure \ref{fig:figure_real_main}. In this setting, models are fine-tuned on a relatively small set of demonstrations. The results demonstrate that the proposed model possesses robust semantic understanding and dexterous action capabilities after training. Overall, AVA-VLA achieves the highest average performance compared to baseline approaches, confirming its real-world applicability. \textit{We visualize the execution trajectories for these tasks in Appendix \ref{sec:app_exp}.}

\subsection{Ablation Studies}\label{sec:ablation}
\noindent\textbf{Model Backbones.~}
%\paragraph{Model Backbones}
To validate the effectiveness of the proposed framework, following \cite{wang2025vla}, we compare three kinds of backbones: The OpenVLA-7B backbone \cite{openvla} pre-trained on robotic data, the prismatic VLM trained on LLaMA2-7B \cite{touvron2023llama}, and the prismatic VLM \cite{karamcheti2024prismatic} trained on Qwen2.5-0.5B \cite{team2024qwen2}. The last two are different-scale backbones without pre-training on robotic data. We compare the proposed method against the standard OpenVLA-OFT method on the LIBERO-Long task suite in the single task setting. Results reported in Table \ref{tab:abl_1} show that our method improves performance across different backbones, even on backbones not pre-trained on robotic datasets.

\begin{table}[th]
\vskip -0.05in
\caption{Ablation study on the model backbones. Comparison on the LIBERO-Long task suite in the LIBERO benchmark in terms of success rates (\%). The best results of each model backbone setting are highlighted in \textbf{bold}.} 
\label{tab:abl_1}
\begin{center}
\vskip -0.2in
\resizebox{0.46\textwidth}{!}{
\begin{tabular}{l|cc}
\toprule
Backbones & OpenVLA-OFT & AVA-VLA \\
\midrule
OpenVLA-7B \cite{openvla}& 94.5 & \textbf{96.2} (1.7\% $\uparrow$) \\
LLaMA2-7B \cite{touvron2023llama} & 90.0 & \textbf{92.6} (2.6\% $\uparrow$) \\
Qwen2.5-0.5B \cite{team2024qwen2} & 89.4 & \textbf{90.8} (1.4\% $\uparrow$)  \\
\bottomrule
\end{tabular}}
\end{center}
\vskip -0.15in
\end{table}

\noindent\textbf{AVA Module and State-Based Initialization.~}
%\paragraph{AVA module, and Prior-based initialization.} 
The AVA-VLA framework consists of two components: the state-based initialization strategy and the AVA module. To validate their individual effectiveness, we conduct ablation experiments on the LIBERO benchmark. As shown in Table \ref{tab:abl_2}, the two recurrent-state-driven components are complementary. State-based initialization injects the recurrent state into the action placeholder to preserve temporal context, which is especially beneficial on LIBERO-Long. The AVA module uses the recurrent state to reweight visual tokens and suppress irrelevant visual content, leading to consistent gains across suites. Each component alone improves over OpenVLA-OFT, and their combination achieves the best overall performance. 
%\textit{Additional cross-benchmark ablations on CALVIN ``ABC$\to$D" are provided in the Appendix \ref{sec:app_add}.}

\begin{table*}[t]
% \small
\caption{Ablation study on the two key components in the AVA-VLA framework. The results on LIBERO in terms of success rates (\%) under the ``one policy for all 4 suites" setting are reported. The best results in each column are highlighted in \textbf{bold}.
}
\vskip -0.2in
\label{tab:abl_2}
\begin{center}
\resizebox{0.98\textwidth}{!}{
\begin{tabular}{l|cccc|c}
\toprule
Method & Spatial SR (\%) & Object SR (\%) & Goal SR (\%) & Long SR (\%) & Average SR (\%)\\ \midrule
OpenVLA-OFT & 97.7 & 98.0 & 96.1 & 95.3 & 96.8 \\
AVA-VLA (State-based initialization) & 97.2 & 98.8 & 96.6 & 97.2 & 97.5 \\
AVA-VLA (AVA module)  & \textbf{97.8} & 98.6 & 97.0 & 96.6 & 97.5 \\
AVA-VLA (AVA module + State-based initialization) & 97.4 & \textbf{99.4} & \textbf{97.4} & \textbf{97.6} & \textbf{98.0} \\

  \bottomrule
\end{tabular}
}
\vskip -0.4in
\end{center}
\end{table*}

\begin{figure*}[!htb]
\vskip -0.15in
\centering
\includegraphics[width=0.97\linewidth]{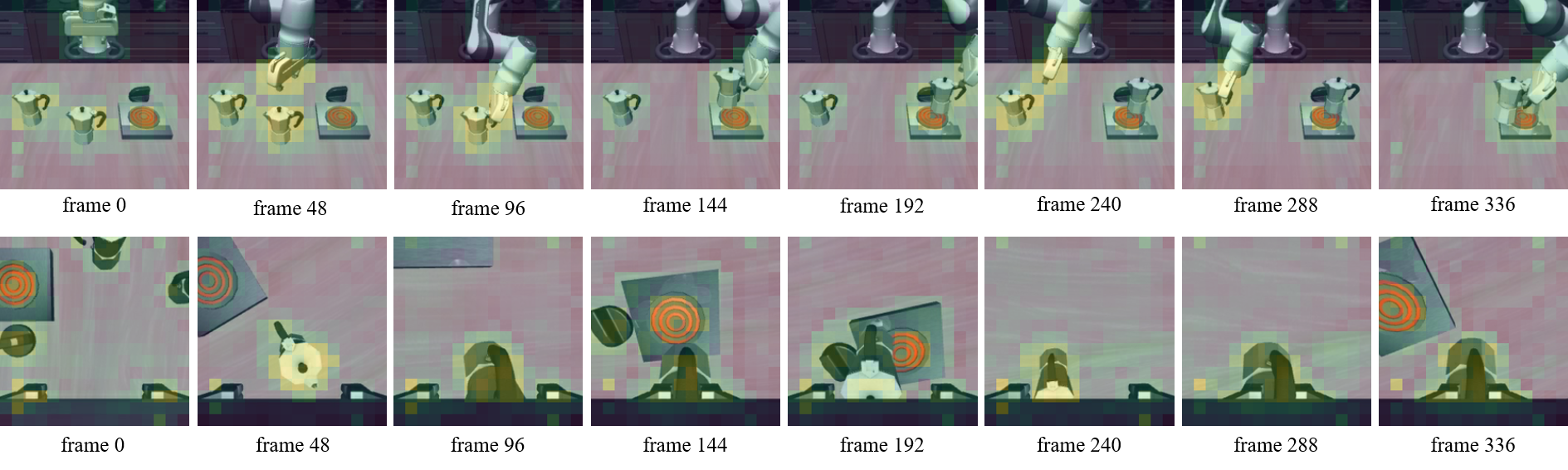}
\vskip -0.1in
\caption{Visual dynamics. The evolution of soft weights during the task ``put both moka pots on the stove" from two viewpoints.}
\label{fig:fig_visul_main}
\vskip -0.15in
\end{figure*}

\begin{table}[t]
% \small
\caption{Study on the visual token pruning with different pruning ratios. The results on LIBERO in terms of success rates (\%) under the ``one policy for all 4 suites" setting are reported.
}
\vskip -0.2in
\label{tab:pruning}
\begin{center}
\resizebox{0.46\textwidth}{!}{
\begin{tabular}{l|cccc|c}
\toprule
Pruning & Spatial  & Object  & Goal  & Long  & Avg. \\
Ratio & SR (\%) & SR (\%) & SR (\%) & SR (\%) & SR (\%)  \\
\midrule
0\% & 97.4 & 99.4 & 97.4 & 97.6 & 98.0\\
50\% & 97.2 & 99.4 & 97.2 & 95.2 & 97.3\\
60\%  & 97.6 & 99.4 & 97.0 & 95.0 & 97.3 \\
70\% & 97.4 & 99.2 & 98.0 & 94.6 & 97.3 \\
80\% & 96.8 & 98.2 & 96.2 & 92.8 & 96.0 \\
90\% & 94.2 & 97.8 & 94.2 & 89.2 & 93.9 \\
\bottomrule
\end{tabular}
}
\end{center}
\vskip -0.4in
\end{table}

\subsection{Analysis}\label{sec:analysis}
\noindent\textbf{Qualitative Visualization.~}
%\paragraph{Qualitative Visualization.}
To investigate the AVA module's focus during task execution, we visualize the soft weights $\boldsymbol{\omega}^t$ across visual tokens during inference. As illustrated in Figure \ref{fig:fig_visul_main}, the attention weights consistently concentrate on the robotic arm's contact regions and the target objects. This selective focus demonstrates the module's ability to identify task-relevant visual features, validating its effectiveness. Furthermore, a direct comparison in Figure \ref{fig:figure_1} reveals that while the vanilla OpenVLA-OFT baseline fails to localize the task-relevant region across viewpoints, AVA-VLA maintains a robust and spatially consistent focus by effectively leveraging historical context.

\noindent\textbf{Visual Token Reduction.~}
%\paragraph{Visual Token Reduction.} 
The proposed AVA module has a potential application in visual token reduction. Although visual token pruning causes the model to lose some visual information, it reduces the model's computational cost and is beneficial for efficient inference \cite{xu2025vla, wang2025unifiedvisionlanguageactionmodel, wang2025specprune, jiang2025better}. To validate that our AVA module effectively prioritizes task-relevant information, we apply a direct ranking strategy to prune visual tokens during inference. Specifically, for a given pruning ratio, we rank all visual tokens by their soft weights and retain only the top-ranked portion corresponding to the desired retention percentage. The results reported in Table \ref{tab:pruning}, demonstrate the robustness of our method: the model suffers only a negligible drop in performance after pruning. Notably, with pruning ratios of 50\%, 60\%, and 70\%, the proposed method continues to outperform the OpenVLA-OFT and maintains performance comparable to the state-of-the-art baselines reported in Table \ref{tab:sota_exps}. The decline in success rate mainly comes from the most challenging LIBERO Long task suite, while the results remain consistent across the other task suites. Even after reducing 90\% of the visual tokens, our method still outperforms many baseline methods listed in Table \ref{tab:sota_exps}. This result further demonstrates the effectiveness of the proposed AVA-VLA framework.

%% file: sec/5_conclusion.tex
\section{Conclusion}

In this paper, we reformulate robot manipulation from a POMDP perspective and propose AVA-VLA, a novel vision-language-action framework for temporally grounded decision-making. Unlike prior VLA models that process each frame independently, our method introduces a recurrent state to approximate the agent’s belief, and builds an Active Visual Attention module to dynamically modulate visual processing of the current observation. In this way, AVA-VLA can actively suppress irrelevant information and focus on task-critical visual features based on historical context.
Extensive experiments demonstrate the superiority of AVA-VLA, achieving state-of-the-art performance across multiple robot simulation benchmarks, including LIBERO and CALVIN, and transferring effectively to diverse real-world robotic tasks. These results highlight the value of recurrent state modeling and history-aware visual processing for robotic sequential decision-making.
% Future work will further improve robustness to long-horizon error accumulation in the recurrent state.

%% file: sec/X_suppl.tex
\clearpage
\setcounter{page}{1}
\maketitlesupplementary
\appendix
\section{Implementation Details}\label{sec:app_imp}
We report the implementation details of our proposed AVA-VLA framework based on the OpenVLA-OFT architecture, and the training details of all experiments.

\noindent\textbf{Base OpenVLA-OFT architecture.~} Our main experiments are based on the OpenVLA-OFT architecture. It integrates a shared SigLIP-DINOv2 backbone for multi-image processing, a Llama-2 7B language model, a 3-layer MLP projector with GELU activation for mapping visual features into the language embedding space, a 2-layer MLP with GELU activation for projecting robot proprioceptive state to the language embedding space, and a 4-layer MLP with ReLU activation for continuous action generation. Distinct from the standard OpenVLA, this architecture replaces causal attention with bidirectional attention to enable parallel decoding, outputting chunks of $\mathrm{L}_{c}$ actions at each timestep.

\noindent\textbf{AVA-VLA framework modifications.~} Our main experiments introduce the following modifications for deploying the AVA-VLA framework on the OpenVLA-OFT foundation model: 1) a 2-layer MLP with SiLU activation for mapping hidden state to the aforementioned recurrent state, 2) three 2-layer MLPs with SiLU activation for mapping visual features, instruction feature, and recurrent state from $d$-dimension to $d'$-dimension, respectively, 3) a feature-wise linear modulation (FiLM) to condition the visual features on the language instruction, 4) a cross-attention layer, a self-attention layer, a FFN, and a linear layer with Softmax activation for predicting the logits for enhancing or weakening each visual token, 5) replacement of the empty placeholder embedding with the recurrent state, 6) modification of the final attention weight matrix based on calculated soft weights vector from the AVA module. 

The proposed AVA-VLA framework introduces only lightweight additional components on top of OpenVLA-OFT. In total, these AVA-related modules add fewer than 50M parameters, accounting for less than 1\% of the full model size. Therefore, the parameter and compute overhead introduced by our modifications are negligible relative to the backbone model.

\noindent\textbf{Training Details.~} For the experiments on the LIBERO benchmark, we use their corresponding official OpenVLA-OFT checkpoints. 
To fine-tune the AVA-VLA model, we apply LoRA \cite{hu2022lora} with a rank of 32 to the LLM backbone, vision encoder, action head and proprioceptive projector, while fully optimizing the proposed AVA mechanism. We set the observation sequence length $K=4$. For efficient training, the gradient is detached between the second and the third timestep. Hyperparameters are set as follows: $\lambda=1.0$, $c=0.6$, $\gamma=[1.9, 0.1]$. The action chunk size is set to $\mathrm{L}_c=8$. The batch size is set to 64. The model is trained for 40,000 gradient steps with an initial learning rate of 5e-4, which includes a warm-up phase by 10\% of the value for stability. Additionally, a cosine learning rate scheduler and a maximum gradient norm of 1.0 is used. For the ablation study on the model backbone in Section \ref{sec:ablation}, OpenVLA-OFT models follow standard implementation details \cite{openvla-oft} with varying initializations, and the AVA-VLA models are trained with implementation details described above using the corresponding OpenVLA-OFT models as initialization.

For the CALVIN benchmark, we train the base OpenVLA-OFT architecture following standard settings in \cite{openvla-oft} using official checkpoints. The AVA-VLA model is trained using the same configuration as the LIBERO benchmark, with the exception of setting $c=0.2$ to account for the smaller region of interest.

For Mobile ALOHA real-world experiments, inputs include one third-person and two wrist-mounted camera images (left wrist + right wrist), we provide the implementation details of the three comparison methods. 
For the UniVLA baseline, following \cite{univla}, we fine-tune the pre-trained checkpoint using the recommended configuration. We employ the latent action decoder on primary images to obtain latent action supervision.  We incorporate proprioceptive states into the action head and integrate dual wrist camera feeds as additional LLM inputs. The action chunk size is set to 25. The model is fine-tuned for 30,000 steps with a learning rate of 3.5e-4, which is decayed to 3.5e-5 after 24,000 steps.
For the OpenVLA-OFT baseline, following \cite{openvla-oft}, we use the official OpenVLA checkpoints as initialization, and apply the LoRA technique with a rank of 32 to the vision encoder and LLM backbone, while the action head and proprioceptive projector are fully optimized. The action chunk size is set to $\mathrm{L}_c=25$. The model is trained for 100,000 gradient steps with an initial learning rate of 5e-4. The learning rate is decayed to 5e-5 after 50,000 steps. The batch size is set to 64. 
For the AVA-VLA model, we utilize the trained OpenVLA-OFT model as initialization, and train the model for 20,000 gradient steps following our LIBERO hyperparameter settings, maintaining an action chunk size of $\mathrm{L}_c=25$.

\noindent\textbf{Training initialization and continued training.~} In our main training recipe, AVA-VLA is initialized from a post-trained OpenVLA-OFT checkpoint. This design is intended to provide a better recurrent-state initialization and improve optimization efficiency of the proposed modules, rather than to gain performance simply by extending training. To verify this explicitly, we additionally compare OpenVLA-OFT and AVA-VLA under matched training settings in Appendix \ref{sec:app_add}.

% \noindent\textbf{Token Reduction Details.} In Section \ref{sec:analysis}, we conduct an ablation study on visual token reduction. We introduce the formal description for this experiment. For given $\boldsymbol{\omega}^t$ and pruning rate $\kappa$, we first 

\begin{table*}[!th]
% \vspace{-10pt}
\centering
\caption{\textbf{Model performance under different perturbations in the LIBERO+ benchmark.} For each column, the average task success rate (\%) of four task suites (LIBERO-Spatial, LIBERO-Object, LIBERO-Goal, and LIBERO-Long) under the given perturbation type is reported. The last column reports the average task success rate over seven perturbation types. The best results in each column of each group are highlighted in \textbf{bold}.}
\resizebox{0.90\linewidth}{!}{%
% \small
\begin{tabular}{l|ccccccc|c}\toprule
Method&Camera &Robot &Language &Light &Background &Noise &Layout &Average \\
\midrule
\multicolumn{9}{c}{\textit{One policy for all 4 suites}} \\
WorldVLA \cite{cen2025worldvla} & 0.1 & \textbf{27.9} & 41.6 & 43.7 & 17.1 & 10.9 & 38.0 & 25.0 \\
$\pi_0$ \cite{pi0} & 13.8 & 6.0 & 58.8 & 85.0 & 81.4 & \textbf{79.0} & 68.9 & 53.6 \\
$\pi_0$-FAST \cite{pi0-fast} & \textbf{65.1} & 21.6 & 61.0 & 73.2 & 73.2 & 74.4 & 68.8 & 61.6 \\
OpenVLA-OFT \cite{openvla-oft} & 55.6 & 21.7 & 81.0 & 92.7 & \textbf{91.0} & 78.6 & 68.7 & 67.9 \\
\midrule
AVA-VLA (Ours) & 55.5 & 25.9 & \textbf{85.6} & \textbf{95.5} & 88.9 & 78.0 & \textbf{74.1} & \textbf{70.1} \\
\midrule[1.2pt]
\multicolumn{9}{c}{\textit{One policy per suite}} \\
OpenVLA \cite{openvla} & 0.8 & 3.5 & 23.0 & 8.1 & 34.8 & 15.2 & 28.5 & 15.6 \\
NORA \cite{hung2025nora} & 2.2 & 37.0 & 65.1 & 45.7 & 58.6 & 12.8 & 62.1 & 39.0 \\
UniVLA \cite{univla} & 1.8 & \textbf{46.2} & 69.6 & 69.0 & 81.0 & 21.2 & 31.9 & 42.9 \\
OpenVLA-OFT \cite{openvla-oft} & 56.4 & 31.9 & 79.5 & 88.7 & 93.3 & 75.8 & 74.2 & 69.6 \\
RIPT-VLA \cite{tan2025interactive} & 55.2 & 31.2 & 77.6 & 88.4 & 91.6 & 73.5 & 74.2 & 68.4 \\
\midrule
AVA-VLA (Ours) & \textbf{69.4} & 34.9 & \textbf{81.5} & \textbf{97.5} & \textbf{94.1} & \textbf{79.1} & \textbf{78.3} & \textbf{74.7} \\
\bottomrule
\end{tabular}
}
\vspace{-10pt}
\label{tab:LIBERO+}
\end{table*}

\begin{table}[!tbp]
    \centering
	\caption{
    \textbf{Comparison under matched training settings.} The results on the LIBERO benchmark in terms of success rates (\%) under the ``one policy for all 4 suites" setting are reported. Both OpenVLA-OFT and AVA-VLA are initialized from the same pretrained OpenVLA checkpoint and trained with 100K gradient steps in a batch size of 256. The best results in each column are highlighted in \textbf{bold}.}
    \vskip -0.1in
    \resizebox{\linewidth}{!}{
\begin{tabular}{l|cccc|c}
\toprule
\multirow{2}{*}{Method} & Spatial  & Object  & Goal  & Long  & Avg. \\
 & SR (\%) & SR (\%) & SR (\%) & SR (\%) & SR (\%)  \\
\midrule
OpenVLA-OFT  & 97.0 & 98.8 & 96.0 & 95.2 & 96.8\\
AVA-VLA & \textbf{98.4} & \textbf{99.4} & \textbf{98.4} & \textbf{96.8} & \textbf{98.3} \\
\bottomrule
\end{tabular}}
\vskip -0.1in
\label{tab:matched_settings}
\end{table}

\begin{table}[!tbp]
    \centering
	\caption{
    \textbf{Ablation study of the loss design on the LIBERO benchmark.} The results in terms of success rates (\%) under the ``one policy for all 4 suites" setting are reported. We remove the L2 penalty regularizer $L_{\omega}$ while keeping all other training settings unchanged. The best results in each column are highlighted in \textbf{bold}.}
    \vskip -0.1in
    \resizebox{\linewidth}{!}{
\begin{tabular}{l|cccc|c}
\toprule
\multirow{2}{*}{Method} & Spatial  & Object  & Goal  & Long  & Avg. \\
 & SR (\%) & SR (\%) & SR (\%) & SR (\%) & SR (\%)  \\
\midrule
AVA-VLA & \textbf{97.4} & \textbf{99.4} & \textbf{97.4} & \textbf{97.6} & \textbf{98.0} \\
AVA-VLA w/o $L_{\omega}$ & 97.4 & 98.8 & 97.2 & 96.4 & 97.5\\
\bottomrule
\end{tabular}}
\vskip -0.2in
\label{tab:ablation_loss_design}
\end{table}

\begin{table}[!tbp]
    \centering
	\caption{
    \textbf{Ablation study of the two modules on the CALVIN ABC$\to$D benchmark.} "+init" denotes enabling state-based initialization only, and "+ava" denotes enabling the AVA module only. The results are reported in terms of success rates (\%) and average length. The best results in each column are highlighted in \textbf{bold}.}\label{tab:abl_2_calvin}
    \vskip -0.1in
    \resizebox{0.5\textwidth}{!}{
\begin{tabular}{l|ccccc|c}
\toprule
    CALVIN   & \multicolumn{5}{c|}{~ Task completed in a row $\uparrow$} & Avg. len \\
     ABC$\to$D & 1 & 2 & 3 & 4 & 5 &$\uparrow$  \\
\midrule
OpenVLA-OFT & 96.9 & 92.0 & 85.7 & 80.4 & 72.9 & 4.28\\
+init & 99.5 & 96.9 & 93.4 & \textbf{90.0} & 83.6 & 4.63 \\
+ava & 99.1 & 96.5 & 93.1 & 89.2 & 82.7 & 4.61 \\
AVA-VLA  & \textbf{99.6} & \textbf{97.6} & \textbf{94.1} & 89.9 & \textbf{84.1} & \textbf{4.65} \\
  \bottomrule
\end{tabular}}
\vskip -0.2in
\end{table}

\section{Real-World Experiment Details}\label{sec:app_exp}
\begin{figure}[!h]
\centering
\includegraphics[width=0.9\linewidth]{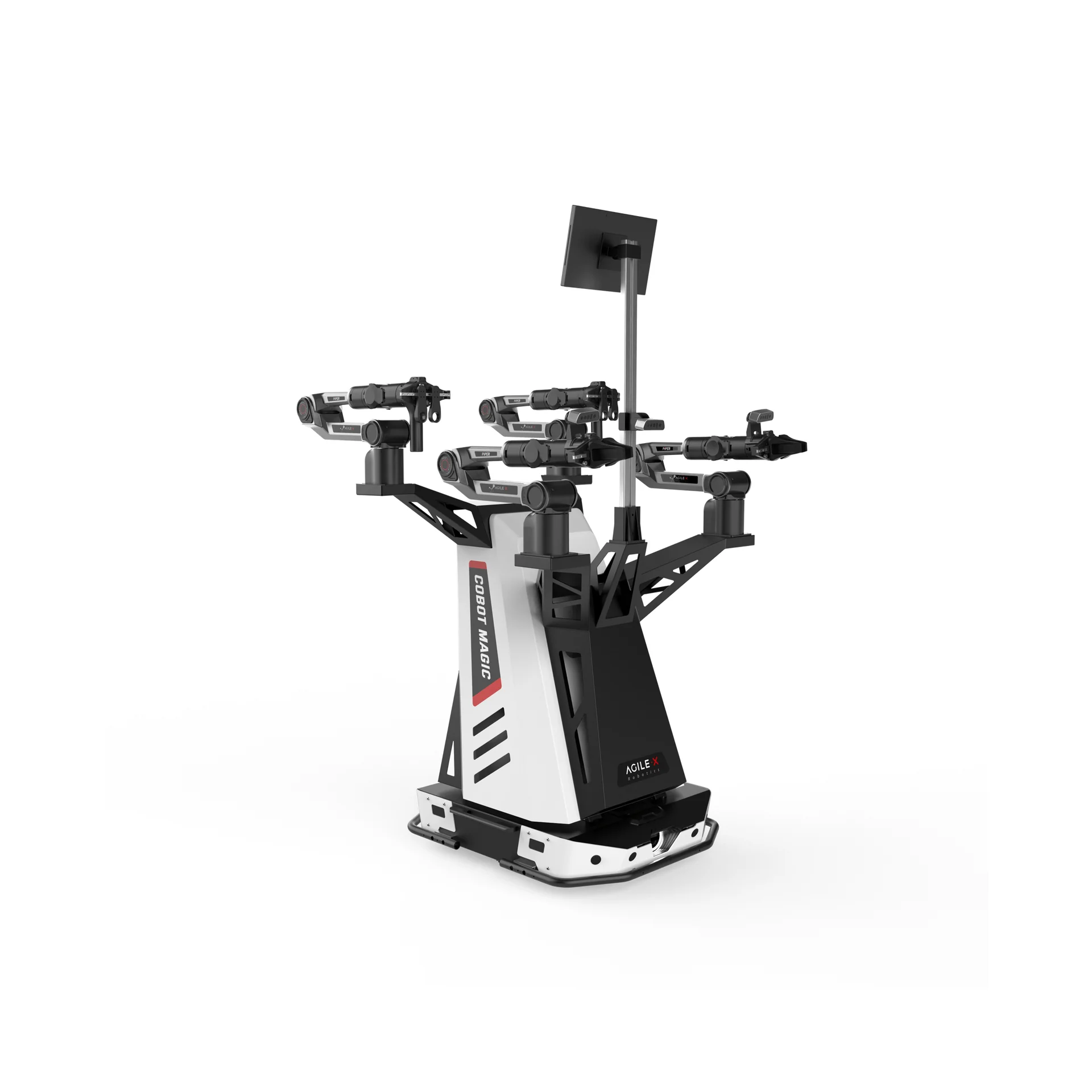}
\vskip -0.2in
\caption{\textbf{AgileX Cobot Magic platforms.}}
\label{fig:real_platform}
\vskip -0.1in
\end{figure}

%仿真数据集细节，真机细节
In this section, we report the additional details of the Mobile ALOHA real-world experiments, including the task suites and execution trajectories.
% 真机平台描述% https://global.agilex.ai/products/cobot-magic
We adopt AgileX Cobot Magic platforms: Based on Stanford’s Mobile ALOHA project \footnote{``https://global.agilex.ai/products/cobot-magic"}, this platform includes a differential-drive AGV base Tracer, dual-arm manipulators, and RGB-D sensors. A platform demonstration can be seen in Figure \ref{fig:real_platform}.

\subsection{Real-World Task Suites}
We introduce the detailed specifications for each task suite in our Mobile ALOHA real-world experiments:

\noindent\textbf{Pick and Place}
\begin{itemize}
    \item Instructions: ``put X into bucket".
    \item Task: Place the bucket in the center and put the simulated toy objects of which the instruction has given (yellow banana, green pepper, purple eggplant) into the bucket.
    \item Dataset:  450 demonstrations (150 per target).
    \item Episode length: 700 timesteps (28 seconds).
    \item Evaluation: 30 trials (10 for each).
\end{itemize}

\noindent\textbf{Sequenced Instruction Understanding}
\begin{itemize}
    \item Instruction: ``stack tower of hanoi".
    \item Task: Stack the medium tower on top of the large one first, and then stack the small one on top of the medium one.
    \item Dataset:  60 demonstrations (10 per formulation).
    \item Episode length: 600 timesteps (24 seconds).
    \item Evaluation: 24 trials (4 for each).
\end{itemize}

\noindent\textbf{Flexible Object Folding}
\begin{itemize}
    \item Instruction: ``fold towel twice".
    \item Task: First fold the towel vertically, then fold horizontally, and finally flatten it.
    \item Dataset:  30 demonstrations.
    \item Episode length: 900 timesteps (36 seconds).
    \item Evaluation: 24 trials.
\end{itemize}

\noindent\textbf{Dexterous Action}
\begin{itemize}
    \item Instructions: ``scoop X into bowl".
    \item Task: Move the bowl to the center of vision, pick up and use the shovel to scoop up different objects (corn, sesame, sunflower seeds) and transfer them into the bowl.
    \item Dataset: 60 demonstrations (20 of each small object).
    \item Episode length: 1000 timesteps (40 seconds).
    \item Evaluation: 24 trials (8 for each).
\end{itemize}

\subsection{Execution Trajectories}

\begin{figure*}[!hb]
\centering
\includegraphics[width=0.90\linewidth]{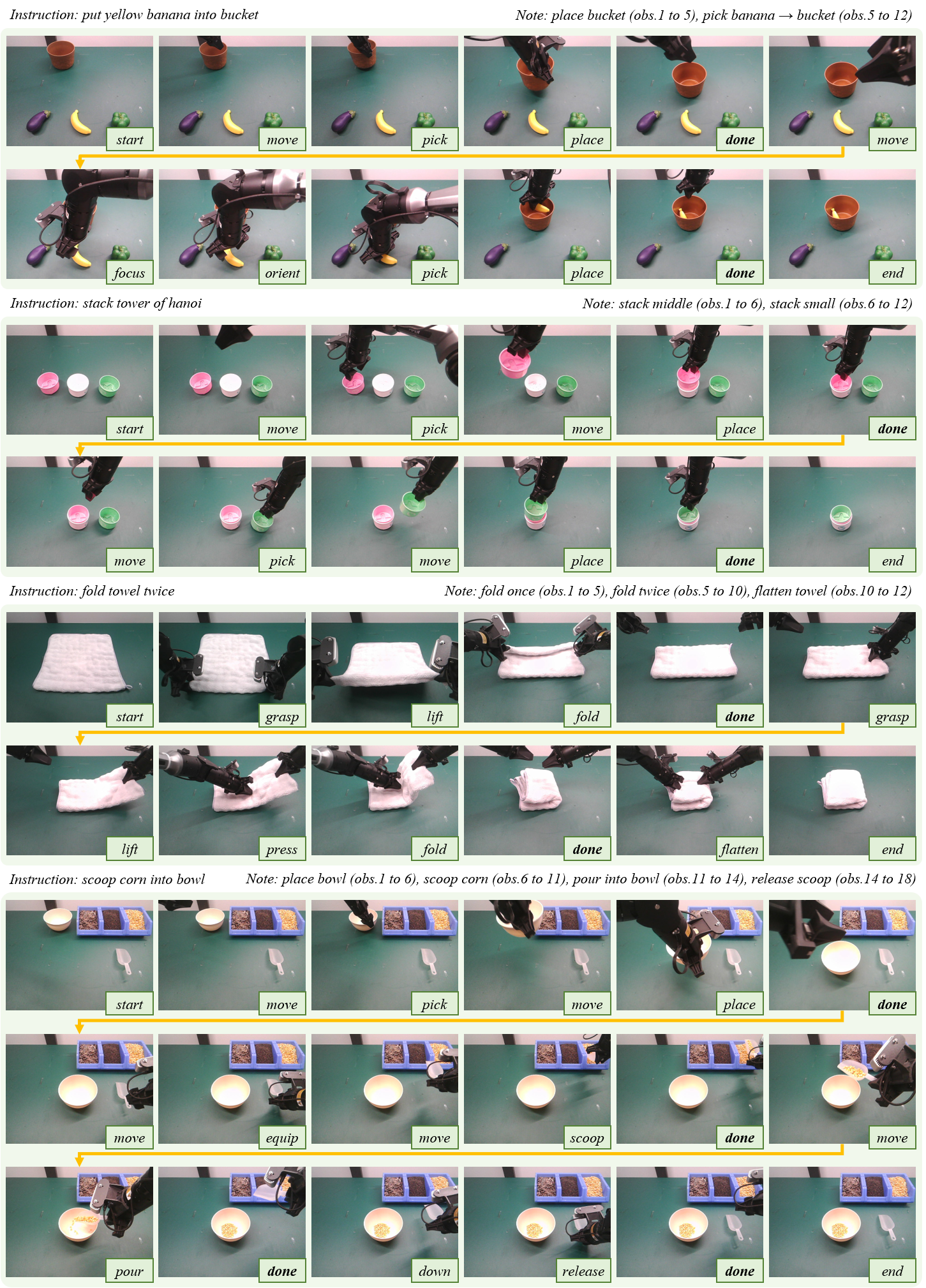}
\vskip -0.01in
\caption{\textbf{Real-world task execution.} Key observations from four long-horizon manipulation tasks.}
\label{fig:execution_real}
\vskip -0.1in
\end{figure*}

We provide the execution trajectories of the four real-world task suites in Figure \ref{fig:execution_real}. The proposed AVA-VLA method can perform various tasks in real-world scenarios.

\section{Additional Discussions}
The proposed AVA-VLA framework is different from recent memory-augmented VLA models, such as MemoryVLA \cite{shi2025memoryvla}. MemoryVLA relies on an explicit, large-scale memory bank for retrieval-based feature augmentation, while AVA-VLA adopts a formal POMDP formulation to compress historical interactions into a compact, implicit recurrent state. Moreover, MemoryVLA focuses on augmenting current features with historical tokens, while our AVA-VLA method utilizes the recurrent state to dynamically modulate and prune visual tokens at the input level, enabling active visual perception that focuses on task-relevant regions.

The significance of temporal modeling and memory mechanisms is well-established across various fields, such as Vision-Language Navigation (VLN) \cite{hong2021vln, zhang2024navid, zhang2025safevla} and Reinforcement Learning \cite{ni2021recurrent, hausknecht2015deep}. Unlike the explicit memory-bank architectures in VLN-BERT \cite{hong2021vln} and SafeVLA \cite{zhang2025safevla} or the LSTM-based aggregation in NaVid \cite{zhang2024navid}, our approach explicitly incorporates a recurrent state based on POMDP to enhance visual representations in a simple yet effective manner. Furthermore, while conceptually related to POMDP-inspired RL algorithms such as Recurrent-PPO \cite{ni2021recurrent} or DRQN \cite{hausknecht2015deep}, AVA-VLA is specifically tailored for VLA tasks, prioritizing visual processing efficiency and the focus on task-relevant features over general policy stability.

\section{Additional Experimental Results}\label{sec:app_add}

In this section, we provide additional evidence for three aspects of AVA-VLA: (i) the gain is not explained by extra training compute, (ii) the proposed design generalizes across benchmarks and controlled perturbations, and (iii) the learned attention is interpretable.

\subsection{Comparison Under Matched Training Settings}
To rule out the confounding effect of additional training compute, we conduct an additional matched-setting comparison (Table \ref{tab:matched_settings}), where OpenVLA-OFT and AVA-VLA are initialized from the same pretrained OpenVLA checkpoint and trained under identical settings, including the same equivalent batch size and the same number of optimization steps. Under this controlled setup, AVA-VLA consistently outperforms OpenVLA-OFT (and its performance is even better than that reported in Table \ref{tab:sota_exps}), indicating that the gain is not explained by a larger training compute alone. Combined with the small parameter overhead of AVA-VLA (\textless{50M}, \textless{1\%} of the full model), these results suggest that the performance gain mainly comes from the proposed architectural synergy between recurrent-state initialization and active visual attention.

\subsection{Module Ablation on CALVIN Benchmark}
To further evaluate whether the effects of the two modules generalize beyond LIBERO, we conduct the same ablation study on the CALVIN benchmark. As shown in Table \ref{tab:abl_2_calvin}, both state-based initialization and AVA consistently improve over OpenVLA-OFT, while their combination achieves the best overall results. Importantly, the gains become more pronounced as the task horizon increases. These results support the same conclusion as Table \ref{tab:abl_2} in the main paper: state-based initialization preserves temporal belief across steps, AVA refines perception by suppressing irrelevant visual content, and the two components are complementary, especially in long-horizon settings.

\subsection{Robustness on LIBERO+ Benchmark}
The LIBERO+ \cite{fei2025libero} benchmark enables us to perform a systematic vulnerability analysis by introducing controlled perturbations across seven dimensions: camera viewpoints (change the viewpoint/pose and field-of-view of the third-person camera), robot initial states (change the manipulator’s initial pose), language instructions (rewrite task instructions to increase linguistic richness and complexity), light conditions (vary illumination intensity, direction, color, and shadow patterns), background textures (modify table/scene textures and materials), sensor noise (inject photometric distortions into input images), and object layout (add confounding objects and/or shift the target object’s position).

We evaluate the proposed method on the LIBERO+ benchmark using the AVA-VLA models trained on the LIBERO benchmark. We do not use additional data to train these models. The evaluation results of two different settings: ``one policy for all 4 suites" and ``one policy per suite" are reported in Table \ref{tab:LIBERO+}. The results of baselines in LIBERO+ benchmarks are based on original references \cite{fei2025libero}. The results show that the proposed AVA-VLA method achieves the best total results over the seven perturbation types on two different settings, demonstrating the superiority of the proposed framework. Notably, the AVA-VLA model exhibits strong robustness under the Light and the Layout perturbations, further demonstrating that the proposed AVA module helps the model enhance the important visual information and reduce the interference of unimportant parts on prediction, thereby improving the model's robustness under visual interference.

\subsection{Effect of the $L_{\omega}$ Regularizer}
We further analyze the regularization term $L_{\omega}$ introduced in Section \ref{sec:method_train} of the main paper. Specifically, we remove the L2 penalty on the soft attention weights while keeping all other training settings unchanged. Quantitative results on the LIBERO benchmark are reported in Table \ref{tab:ablation_loss_design}. Removing $L_{\omega}$ reduces the average success rate from 98.0\% to 97.5\%, with the most noticeable drop on the LIBERO-Long suite.

\begin{figure*}[!htb]
    \centering
    \includegraphics[width=0.99\linewidth]{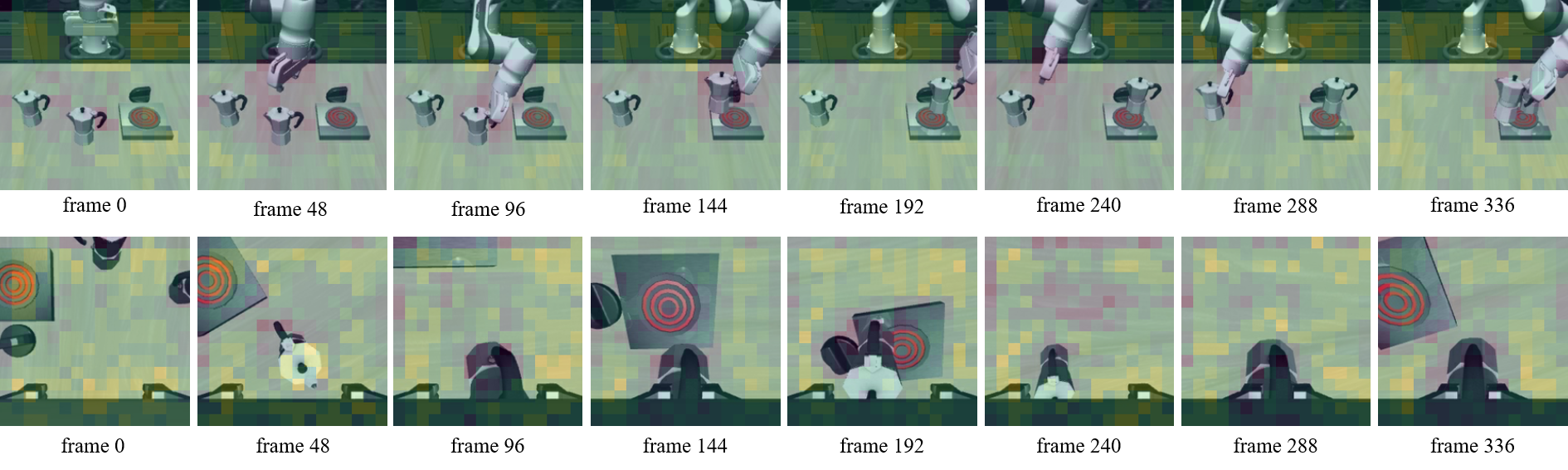}
    \caption{\textbf{Visualization of the soft weights without the regularizer $L_{\omega}$ on LIBERO.} Compared with the full AVA-VLA result shown in Figure \ref{fig:fig_visul_main}, removing $L_{\omega}$ leads to more dispersed attention and increased responses on irrelevant background regions, indicating that the regularizer helps maintain more selective and structurally robust attention masks.}
    \label{fig:ablation_loss_design}
\end{figure*}

We also visualize the effect of removing the regularization term $L_{\omega}$ on the same task instance used in Figure \ref{fig:fig_visul_main} of the main paper. As shown in Figure~\ref{fig:ablation_loss_design}, without $L_{\omega}$, the learned soft attention becomes noticeably more dispersed and allocates more mass to irrelevant background regions. Compared with the full model visualization in Figure \ref{fig:fig_visul_main}, this suggests that $L_{\omega}$ stabilizes the sparsity pattern of AVA and helps the model maintain task-relevant focus over time.

%\section{Supplementary Quantitative Analysis}\label{sec:app_ana}
\subsection{Attention Visualization Across Tasks}
In Section \ref{sec:analysis}, we visualize the soft weights $\boldsymbol{\omega}^t$ related to the corresponding visual tokens during the inference of one example from the LIBERO benchmark. Additionally, we present further visualizations of the soft weights calculated by the AVA module across a broader set of examples to demonstrate the proposed method's consistency.

\begin{figure*}[!htb]
\centering
\includegraphics[width=1\linewidth]{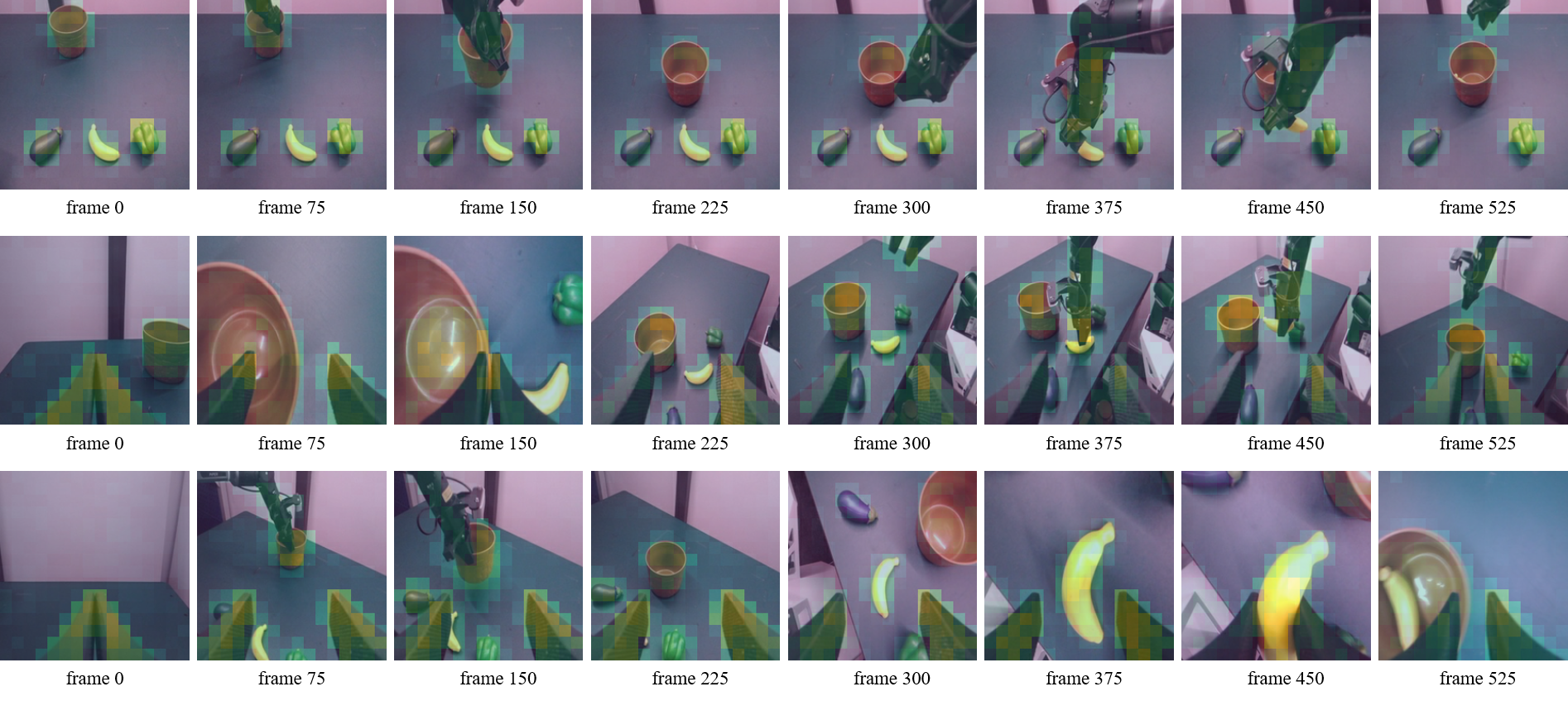}
\vskip -0.01in
\caption{\textbf{Attention dynamics on Mobile ALOHA.} Soft weights for “put yellow banana into bucket” from three viewpoints.}
\label{fig:attention_real_1}
\vskip -0.1in
\end{figure*}

\begin{figure*}[!htb]
\centering
\includegraphics[width=1\linewidth]{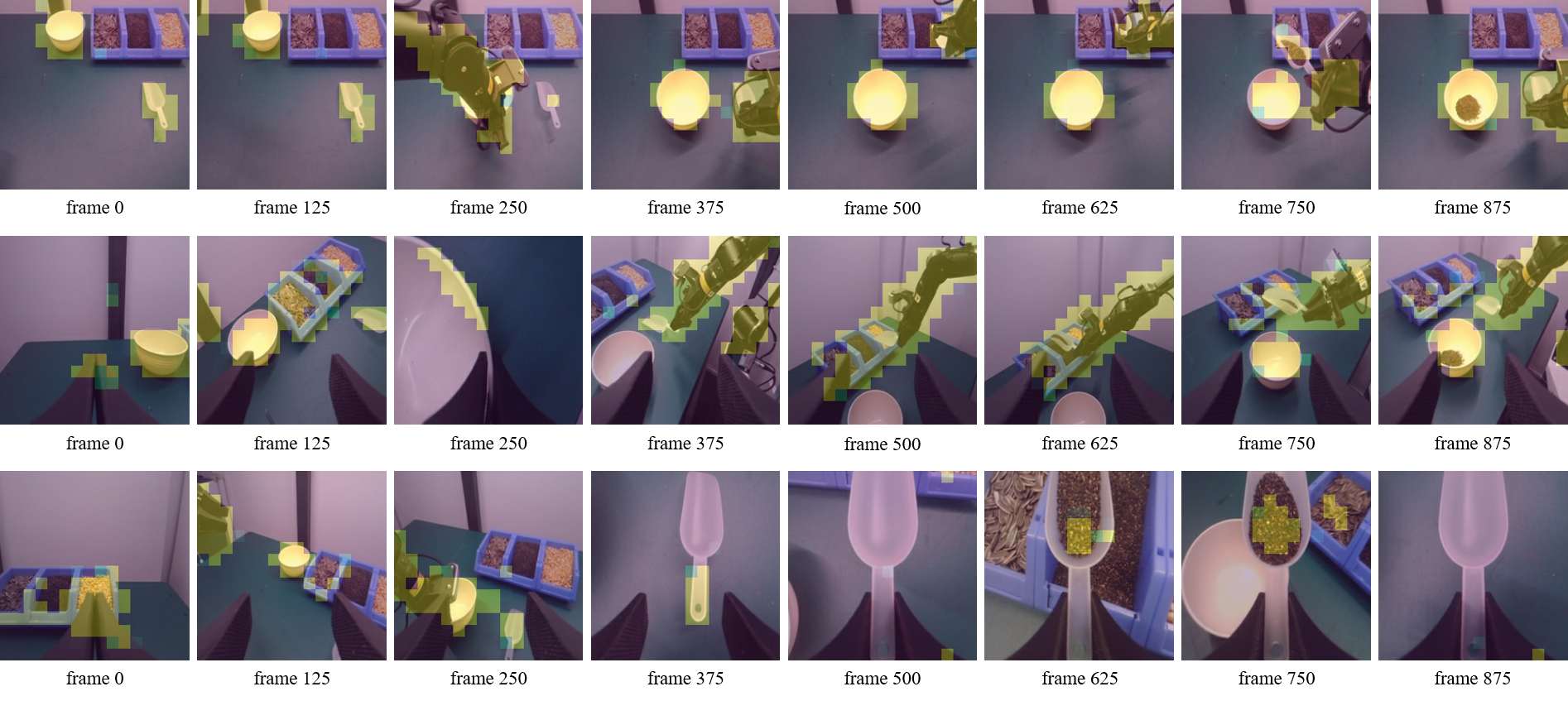}
\vskip -0.01in
\caption{\textbf{Attention dynamics on Mobile ALOHA.} Soft weights for “scoop sesame into bowl” from three viewpoints.}
\label{fig:attention_real_2}
\vskip -0.1in
\end{figure*}

Figure \ref{fig:attention_real_1} and Figure \ref{fig:attention_real_2} illustrate results from Mobile ALOHA real-world experiments, covering two task suites across three viewpoints. The results demonstrate the proposed framework's ability to focus on important visual information. Specifically, in the ``put yellow banana into bucket" task, the model consistently locates and focuses on the objects requiring interaction: the yellow banana and the bucket. Similarly, for the ``scoop sesame into bowl" task, the model accurately pinpoints the interaction target, such as the ladle handle in frame 375 of the right wrist-mounted camera.

\begin{figure*}[!htb]
\centering
\includegraphics[width=0.99\linewidth]{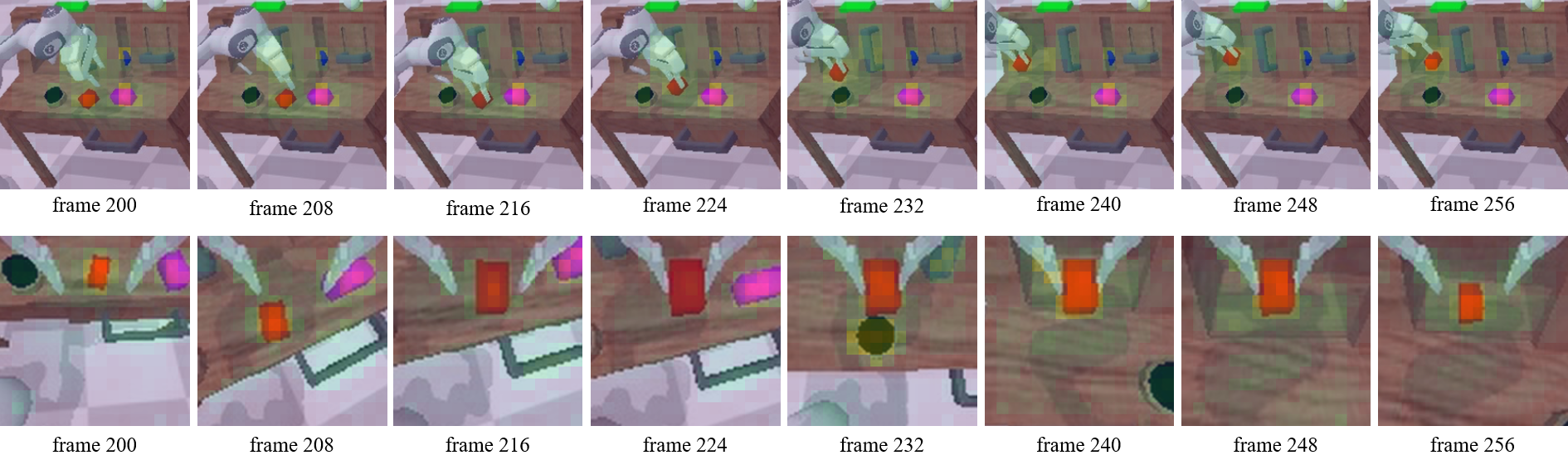}
\vskip -0.01in
\caption{\textbf{Attention dynamics on CALVIN.} Soft weights for the continuous tasks ``Lift red block table" and ``Place in slider" from two viewpoints.}
\label{fig:attention_calvin}
\vskip -0.1in
\end{figure*}

\begin{figure*}[!htb]
\centering
\includegraphics[width=0.99\linewidth]{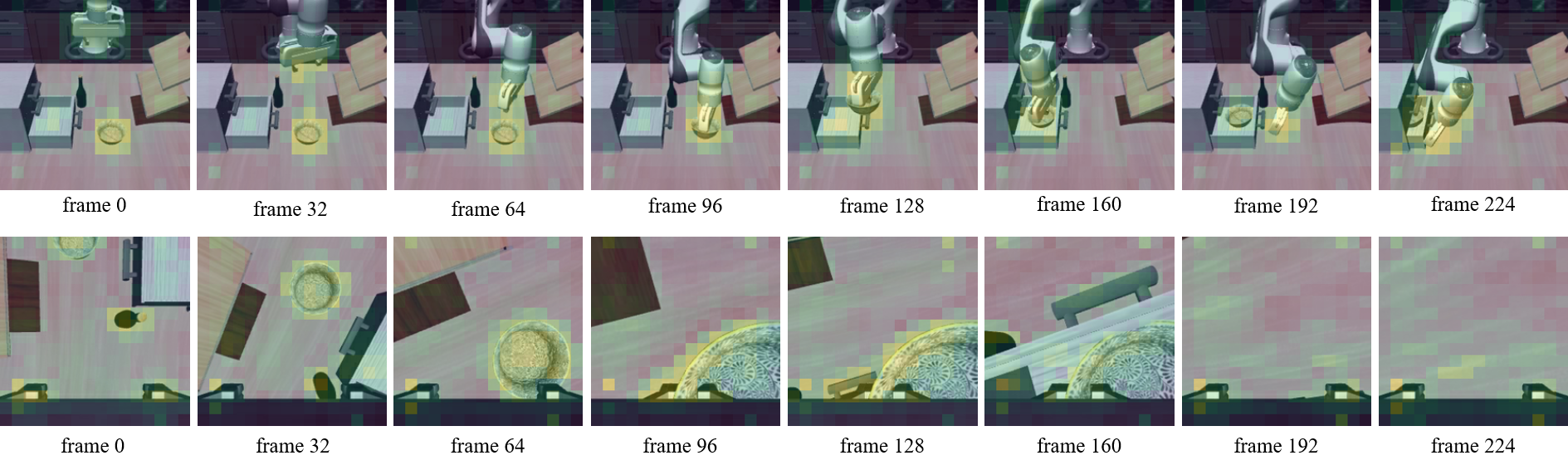}
\vskip -0.01in
\caption{\textbf{Attention dynamics on LIBERO.} Soft weights for ``put the black bowl in the bottom drawer of the cabinet and close it" from two viewpoints.}
\label{fig:attention_libro_2}
\vskip -0.1in
\end{figure*}

\begin{figure*}[!htb]
\centering
\includegraphics[width=0.99\linewidth]{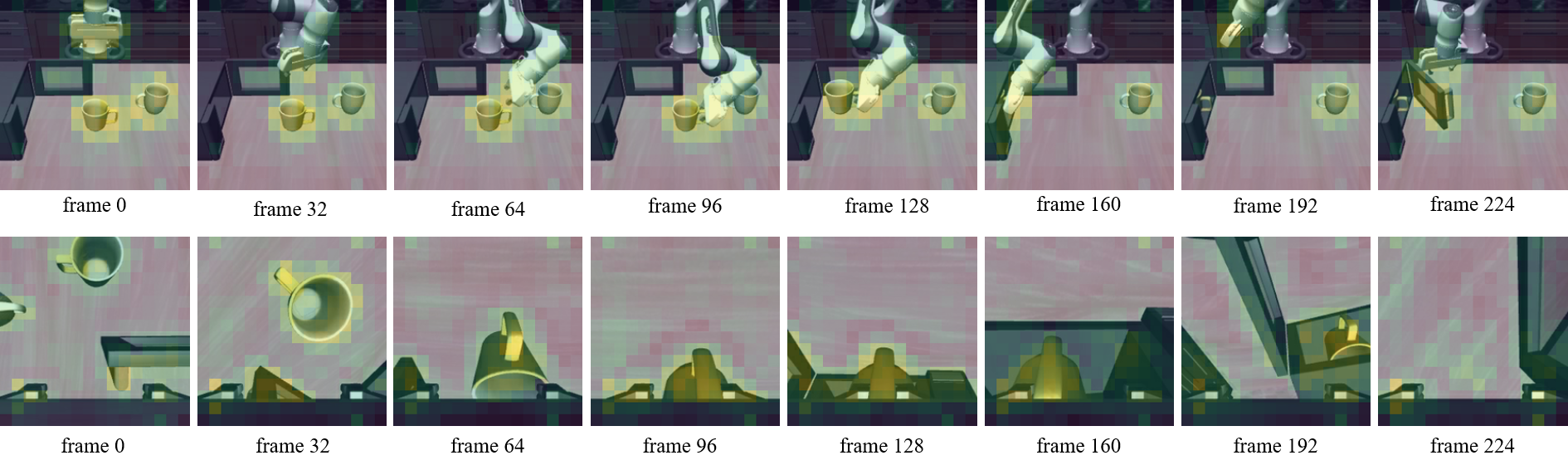}
\vskip -0.01in
\caption{\textbf{Attention dynamics on LIBERO.} Soft weights for ``put the yellow and white mug in the microwave and close it" from two viewpoints.}
\label{fig:attention_libro_3}
\vskip -0.1in
\end{figure*}

Extended visualization results for simulated environments are presented in Figures \ref{fig:attention_calvin}, \ref{fig:attention_libro_2}, and \ref{fig:attention_libro_3}. Figure \ref{fig:attention_calvin} displays the results for the continuous tasks ``Lift red block table" and ``Place in slider" from two viewpoints for the experiment on the CALVIN benchmark. Figure \ref{fig:attention_libro_2} and Figure \ref{fig:attention_libro_3} display the results for two tasks from two viewpoints for the experiment on the LIBERO benchmark, respectively. These additional visualization results on the simulation environments consistently corroborate our findings. The proposed AVA-VLA method can enable the VLA model to effectively enhance the perception of critical visual information while suppressing irrelevant regions, thereby improving the model's performance.

% \subsection{Qualitative Failure Cases}
% We provide qualitative failure cases (Figure \ref{fig:failure_cases}) corresponding to the limitations discussed in Section \ref{sec:analysis}. These cases illustrate how minor perceptual inaccuracies accumulate in the recurrent state, leading to drifted object/contact beliefs and eventual failures in precision-sensitive long-horizon tasks.

\section{Limitations}
\begin{figure*}[!htb]
    \centering
    % --- 第一行图片 ---
    \begin{subfigure}{\linewidth}
        \centering
        \includegraphics[width=0.99\linewidth]{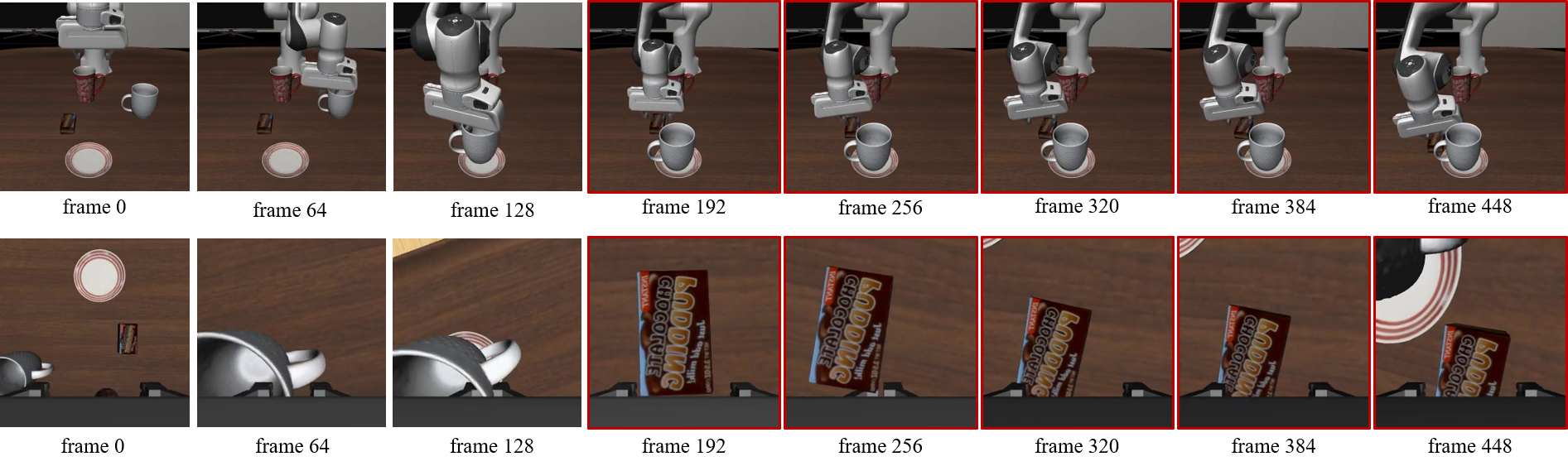}
        % \vskip -0.1in
        \caption{Task: Put the white mug on the plate and put the chocolate pudding to the right of the plate.}
        \label{fig:failure_a}
    \end{subfigure}
    
    \vspace{1em} % 两行之间的垂直间距

    % --- 第二行图片 ---
    \begin{subfigure}{\linewidth}
        \centering
        \includegraphics[width=0.99\linewidth]{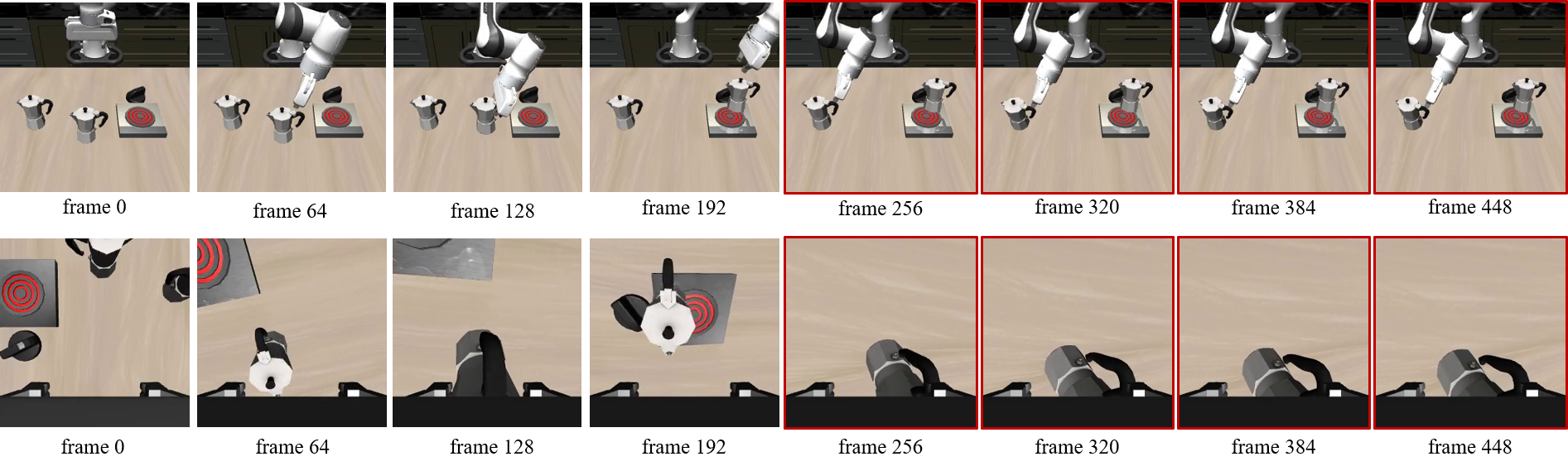}
        % \vskip -0.1in
        \caption{Task: Put both moka pots on the stove.}
        \label{fig:failure_b}
    \end{subfigure}

    \caption{\textbf{Failure cases of AVA-VLA on LIBERO.} (a) The gripper fails to align with the chocolate pudding due to drifted spatial belief. (b) A slight positional deviation prevents the robot from securely grasping the moka pot handle. These cases illustrate how minor perceptual inaccuracies accumulate in the recurrent state, leading to drifted object/contact beliefs and eventual failures in precision-sensitive long-horizon tasks.}
    \label{fig:failure_cases}
\end{figure*}

Despite its strong performance, AVA-VLA still faces a fundamental challenge of POMDP modeling: small perception or state-estimation errors may accumulate over long horizons, gradually leading to belief drift and failures in precision-sensitive manipulation such as grasping or placement. See the provided qualitative failure cases in Figure \ref{fig:failure_cases}. This issue is especially pronounced in long-horizon tasks such as LIBERO-Long, where performance drops more markedly under visual token reduction. A promising direction for future work is to improve the stability of recurrent state propagation, for example, through more robust state-update mechanisms, explicit error-correction strategies, or longer-horizon training schemes that better align the recurrent state with task-relevant environment dynamics.

% \begin{figure*}[!htb]
% \centering
% \includegraphics[width=\linewidth]{fig/app-ED-1.png}
% \vskip -0.01in
% \caption{test for appendix visual example.}
% \label{fig:app_s_1}
% \vskip -0.1in
% \end{figure*}
% \begin{figure*}[!htb]
% \centering
% \includegraphics[width=\linewidth]{fig/app-ED-2.png}
% \vskip -0.01in
% \caption{test for appendix visual example.}
% \label{fig:app_s_2}
% \vskip -0.1in
% \end{figure*}
% \begin{figure*}[!htb]
% \centering
% \includegraphics[width=\linewidth]{fig/app-ED-3.png}
% \vskip -0.01in
% \caption{test for appendix visual example.}
% \label{fig:app_s_3}
% \vskip -0.1in
% \end{figure*}
% \begin{figure*}[!htb]
% \centering
% \includegraphics[width=\linewidth]{fig/app-ED-4.png}
% \vskip -0.01in
% \caption{test for appendix visual example.}
% \label{fig:app_s_4}
% \vskip -0.1in
% \end{figure*}


%% file: main.bib
@String(ICLR = {Int. Conf. Learn. Represent.})

@String(AAAI = {AAAI})

@String(ICLR  = {ICLR})

@article{bjorck2025gr00t,
  title={Gr00t n1: An open foundation model for generalist humanoid robots},
  author={Bjorck, Johan and Casta{\~n}eda, Fernando and Cherniadev, Nikita and Da, Xingye and Ding, Runyu and Fan, Linxi and Fang, Yu and Fox, Dieter and Hu, Fengyuan and Huang, Spencer and others},
  journal={arXiv preprint arXiv:2503.14734},
  year={2025}
}

@inproceedings{liao2018reviving,
  title={Reviving and improving recurrent back-propagation},
  author={Liao, Renjie and Xiong, Yuwen and Fetaya, Ethan and Zhang, Lisa and Yoon, KiJung and Pitkow, Xaq and Urtasun, Raquel and Zemel, Richard},
  booktitle={International conference on machine learning},
  pages={3082--3091},
  year={2018},
  organization={PMLR}
}

@inproceedings{pascanu2013difficulty,
  title={On the difficulty of training recurrent neural networks},
  author={Pascanu, Razvan and Mikolov, Tomas and Bengio, Yoshua},
  booktitle={International conference on machine learning},
  pages={1310--1318},
  year={2013},
  organization={Pmlr}
}

@article{medsker2001recurrent,
  title={Recurrent neural networks},
  author={Medsker, Larry R and Jain, Lakhmi and others},
  journal={Design and applications},
  volume={5},
  number={64-67},
  pages={2},
  year={2001}
}

@inproceedings{zhang2017dynamically,
  title={Dynamically constructed (PO) MDPs for adaptive robot planning},
  author={Zhang, Shiqi and Khandelwal, Piyush and Stone, Peter},
  booktitle={Proceedings of the AAAI conference on artificial intelligence},
  volume={31},
  number={1},
  year={2017}
}

@article{rt1,
  title={Rt-1: Robotics transformer for real-world control at scale},
  author={Brohan, Anthony and Brown, Noah and Carbajal, Justice and Chebotar, Yevgen and Dabis, Joseph and Finn, Chelsea and Gopalakrishnan, Keerthana and Hausman, Karol and Herzog, Alex and Hsu, Jasmine and others},
  journal={arXiv preprint arXiv:2212.06817},
  year={2022}
}

@article{liu2023visual,
  title={Visual instruction tuning},
  author={Liu, Haotian and Li, Chunyuan and Wu, Qingyang and Lee, Yong Jae},
  journal={Advances in neural information processing systems},
  volume={36},
  pages={34892--34916},
  year={2023}
}

@article{belkhale2024rt,
  title={Rt-h: Action hierarchies using language},
  author={Belkhale, Suneel and Ding, Tianli and Xiao, Ted and Sermanet, Pierre and Vuong, Quon and Tompson, Jonathan and Chebotar, Yevgen and Dwibedi, Debidatta and Sadigh, Dorsa},
  journal={arXiv preprint arXiv:2403.01823},
  year={2024}
}

@inproceedings{rt2,
  title={Rt-2: Vision-language-action models transfer web knowledge to robotic control},
  author={Zitkovich, Brianna and Yu, Tianhe and Xu, Sichun and Xu, Peng and Xiao, Ted and Xia, Fei and Wu, Jialin and Wohlhart, Paul and Welker, Stefan and Wahid, Ayzaan and others},
  booktitle={Conference on Robot Learning},
  pages={2165--2183},
  year={2023},
  organization={PMLR}
}

@article{pi0,
  title={$\pi_0$: A Vision-Language-Action Flow Model for General Robot Control},
  author={Black, Kevin and Brown, Noah and Driess, Danny and Esmail, Adnan and Equi, Michael and Finn, Chelsea and Fusai, Niccolo and Groom, Lachy and Hausman, Karol and Ichter, Brian and others},
  journal={arXiv preprint arXiv:2410.24164},
  year={2024}
}

@article{openvla,
  title={Openvla: An open-source vision-language-action model},
  author={Kim, Moo Jin and Pertsch, Karl and Karamcheti, Siddharth and Xiao, Ted and Balakrishna, Ashwin and Nair, Suraj and Rafailov, Rafael and Foster, Ethan and Lam, Grace and Sanketi, Pannag and others},
  journal={arXiv preprint arXiv:2406.09246},
  year={2024}
}

@article{qu2025spatialvla,
  title={Spatialvla: Exploring spatial representations for visual-language-action model},
  author={Qu, Delin and Song, Haoming and Chen, Qizhi and Yao, Yuanqi and Ye, Xinyi and Ding, Yan and Wang, Zhigang and Gu, JiaYuan and Zhao, Bin and Wang, Dong and others},
  journal={arXiv preprint arXiv:2501.15830},
  year={2025}
}

@inproceedings{sun2025lvpruning,
  title={Lvpruning: An effective yet simple language-guided vision token pruning approach for multi-modal large language models},
  author={Sun, Yizheng and Xin, Yanze and Li, Hao and Sun, Jingyuan and Lin, Chenghua and Batista-Navarro, Riza Theresa},
  booktitle={Findings of the Association for Computational Linguistics: NAACL 2025},
  pages={4299--4308},
  year={2025}
}

@article{rao2021dynamicvit,
  title={Dynamicvit: Efficient vision transformers with dynamic token sparsification},
  author={Rao, Yongming and Zhao, Wenliang and Liu, Benlin and Lu, Jiwen and Zhou, Jie and Hsieh, Cho-Jui},
  journal={Advances in neural information processing systems},
  volume={34},
  pages={13937--13949},
  year={2021}
}

@article{lin2024petformer,
  title={Petformer: Long-term time series forecasting via placeholder-enhanced transformer},
  author={Lin, Shengsheng and Lin, Weiwei and Wu, Wentai and Wang, Songbo and Wang, Yongxiang},
  journal={IEEE Transactions on Emerging Topics in Computational Intelligence},
  year={2024},
  publisher={IEEE}
}

@article{tian2024predictive,
  title={Predictive inverse dynamics models are scalable learners for robotic manipulation},
  author={Tian, Yang and Yang, Sizhe and Zeng, Jia and Wang, Ping and Lin, Dahua and Dong, Hao and Pang, Jiangmiao},
  journal={arXiv preprint arXiv:2412.15109},
  year={2024}
}

@article{li2025cogvla,
  title={Cogvla: Cognition-aligned vision-language-action model via instruction-driven routing \& sparsification},
  author={Li, Wei and Zhang, Renshan and Shao, Rui and He, Jie and Nie, Liqiang},
  journal={arXiv preprint arXiv:2508.21046},
  year={2025}
}

@article{cen2025worldvla,
  title={WorldVLA: Towards Autoregressive Action World Model},
  author={Cen, Jun and Yu, Chaohui and Yuan, Hangjie and Jiang, Yuming and Huang, Siteng and Guo, Jiayan and Li, Xin and Song, Yibing and Luo, Hao and Wang, Fan and others},
  journal={arXiv preprint arXiv:2506.21539},
  year={2025}
}

@article{tan2025interactive,
  title={Interactive Post-Training for Vision-Language-Action Models},
  author={Tan, Shuhan and Dou, Kairan and Zhao, Yue and Kr{\"a}henb{\"u}hl, Philipp},
  journal={arXiv preprint arXiv:2505.17016},
  year={2025}
}

@article{wang2025unifiedvisionlanguageactionmodel,
  title={Unified Vision-Language-Action Model},
  author={Yuqi Wang and Xinghang Li and Wenxuan Wang and Junbo Zhang and Yingyan Li and Yuntao Chen and Xinlong Wang and Zhaoxiang Zhang},
  journal={arXiv preprint arXiv:2506.19850},
  year={2025}
}

@article{reuss2025flower,
  title={Flower: Democratizing generalist robot policies with efficient vision-language-action flow policies},
  author={Reuss, Moritz and Zhou, Hongyi and R{\"u}hle, Marcel and Ya{\u{g}}murlu, {\"O}mer Erdin{\c{c}} and Otto, Fabian and Lioutikov, Rudolf},
  journal={arXiv preprint arXiv:2509.04996},
  year={2025}
}

@article{song2025accelerating,
  title={Accelerating vision-language-action model integrated with action chunking via parallel decoding},
  author={Song, Wenxuan and Chen, Jiayi and Ding, Pengxiang and Zhao, Han and Zhao, Wei and Zhong, Zhide and Ge, Zongyuan and Ma, Jun and Li, Haoang},
  journal={arXiv preprint arXiv:2503.02310},
  year={2025}
}

@article{hung2025nora,
  title={Nora: A small open-sourced generalist vision language action model for embodied tasks},
  author={Hung, Chia-Yu and Sun, Qi and Hong, Pengfei and Zadeh, Amir and Li, Chuan and Tan, U and Majumder, Navonil and Poria, Soujanya and others},
  journal={arXiv preprint arXiv:2504.19854},
  year={2025}
}

@article{openvla-oft,
  title={Fine-tuning vision-language-action models: Optimizing speed and success},
  author={Kim, Moo Jin and Finn, Chelsea and Liang, Percy},
  journal={arXiv preprint arXiv:2502.19645},
  year={2025}
}

@article{pi0-fast,
  title={Fast: Efficient action tokenization for vision-language-action models},
  author={Pertsch, Karl and Stachowicz, Kyle and Ichter, Brian and Driess, Danny and Nair, Suraj and Vuong, Quan and Mees, Oier and Finn, Chelsea and Levine, Sergey},
  journal={arXiv preprint arXiv:2501.09747},
  year={2025}
}

@inproceedings{cot-vla,
  title={Cot-vla: Visual chain-of-thought reasoning for vision-language-action models},
  author={Zhao, Qingqing and Lu, Yao and Kim, Moo Jin and Fu, Zipeng and Zhang, Zhuoyang and Wu, Yecheng and Li, Zhaoshuo and Ma, Qianli and Han, Song and Finn, Chelsea and others},
  booktitle={Proceedings of the Computer Vision and Pattern Recognition Conference},
  pages={1702--1713},
  year={2025}
}

@article{univla,
  title={Univla: Learning to act anywhere with task-centric latent actions},
  author={Bu, Qingwen and Yang, Yanting and Cai, Jisong and Gao, Shenyuan and Ren, Guanghui and Yao, Maoqing and Luo, Ping and Li, Hongyang},
  journal={arXiv preprint arXiv:2505.06111},
  year={2025}
}

@inproceedings{tracevla,
  title={TraceVLA: Visual Trace Prompting Enhances Spatial-Temporal Awareness for Generalist Robotic Policies},
  author={Zheng, Ruijie and Liang, Yongyuan and Huang, Shuaiyi and Gao, Jianfeng and Daum{\'e} III, Hal and Kolobov, Andrey and Huang, Furong and Yang, Jianwei},
  booktitle={The Thirteenth International Conference on Learning Representations},
  year={2025}
}

@article{shen2024mome,
  title={Mome: Mixture of multimodal experts for generalist multimodal large language models},
  author={Shen, Leyang and Chen, Gongwei and Shao, Rui and Guan, Weili and Nie, Liqiang},
  journal={Advances in neural information processing systems},
  volume={37},
  pages={42048--42070},
  year={2024}
}

@article{chen2023pali,
  title={Pali-3 vision language models: Smaller, faster, stronger},
  author={Chen, Xi and Wang, Xiao and Beyer, Lucas and Kolesnikov, Alexander and Wu, Jialin and Voigtlaender, Paul and Mustafa, Basil and Goodman, Sebastian and Alabdulmohsin, Ibrahim and Padlewski, Piotr and others},
  journal={arXiv preprint arXiv:2310.09199},
  year={2023}
}

@article{hu2022lora,
  title={Lora: Low-rank adaptation of large language models.},
  author={Hu, Edward J and Shen, Yelong and Wallis, Phillip and Allen-Zhu, Zeyuan and Li, Yuanzhi and Wang, Shean and Wang, Lu and Chen, Weizhu and others},
  journal={ICLR},
  volume={1},
  number={2},
  pages={3},
  year={2022}
}

@article{zhang2024sparsevlm,
  title={Sparsevlm: Visual token sparsification for efficient vision-language model inference},
  author={Zhang, Yuan and Fan, Chun-Kai and Ma, Junpeng and Zheng, Wenzhao and Huang, Tao and Cheng, Kuan and Gudovskiy, Denis and Okuno, Tomoyuki and Nakata, Yohei and Keutzer, Kurt and others},
  journal={arXiv preprint arXiv:2410.04417},
  year={2024}
}

@inproceedings{chen2024image,
  title={An image is worth 1/2 tokens after layer 2: Plug-and-play inference acceleration for large vision-language models},
  author={Chen, Liang and Zhao, Haozhe and Liu, Tianyu and Bai, Shuai and Lin, Junyang and Zhou, Chang and Chang, Baobao},
  booktitle={European Conference on Computer Vision},
  pages={19--35},
  year={2024},
  organization={Springer}
}

@article{smallwood1973optimal,
  title={The optimal control of partially observable Markov processes over a finite horizon},
  author={Smallwood, Richard D and Sondik, Edward J},
  journal={Operations research},
  volume={21},
  number={5},
  pages={1071--1088},
  year={1973},
  publisher={INFORMS}
}

@article{wang2025specprune,
  title={Specprune-vla: Accelerating vision-language-action models via action-aware self-speculative pruning},
  author={Wang, Hanzhen and Xu, Jiaming and Pan, Jiayi and Zhou, Yongkang and Dai, Guohao},
  journal={arXiv preprint arXiv:2509.05614},
  year={2025}
}

@article{li2025sp,
  title={SP-VLA: A Joint Model Scheduling and Token Pruning Approach for VLA Model Acceleration},
  author={Li, Ye and Meng, Yuan and Sun, Zewen and Ji, Kangye and Tang, Chen and Fan, Jiajun and Ma, Xinzhu and Xia, Shutao and Wang, Zhi and Zhu, Wenwu},
  journal={arXiv preprint arXiv:2506.12723},
  year={2025}
}

@article{xu2025vla,
  title={Vla-cache: Towards efficient vision-language-action model via adaptive token caching in robotic manipulation},
  author={Xu, Siyu and Wang, Yunke and Xia, Chenghao and Zhu, Dihao and Huang, Tao and Xu, Chang},
  journal={arXiv preprint arXiv:2502.02175},
  year={2025}
}

@article{jiang2025irl,
  title={Irl-vla: Training an vision-language-action policy via reward world model},
  author={Jiang, Anqing and Gao, Yu and Wang, Yiru and Sun, Zhigang and Wang, Shuo and Heng, Yuwen and Sun, Hao and Tang, Shichen and Zhu, Lijuan and Chai, Jinhao and others},
  journal={arXiv preprint arXiv:2508.06571},
  year={2025}
}

@article{mees2022calvin,
  title={Calvin: A benchmark for language-conditioned policy learning for long-horizon robot manipulation tasks},
  author={Mees, Oier and Hermann, Lukas and Rosete-Beas, Erick and Burgard, Wolfram},
  journal={IEEE Robotics and Automation Letters},
  volume={7},
  number={3},
  pages={7327--7334},
  year={2022},
  publisher={IEEE}
}

@inproceedings{wang2023efficientvlm,
  title={Efficientvlm: Fast and accurate vision-language models via knowledge distillation and modal-adaptive pruning},
  author={Wang, Tiannan and Zhou, Wangchunshu and Zeng, Yan and Zhang, Xinsong},
  booktitle={Findings of the association for computational linguistics: ACL 2023},
  pages={13899--13913},
  year={2023}
}

@article{touvron2023llama,
  title={Llama 2: Open foundation and fine-tuned chat models},
  author={Touvron, Hugo and Martin, Louis and Stone, Kevin and Albert, Peter and Almahairi, Amjad and Babaei, Yasmine and Bashlykov, Nikolay and Batra, Soumya and Bhargava, Prajjwal and Bhosale, Shruti and others},
  journal={arXiv preprint arXiv:2307.09288},
  year={2023}
}

@article{team2024qwen2,
  title={Qwen2 technical report},
  author={Team, Qwen and others},
  journal={arXiv preprint arXiv:2407.10671},
  volume={2},
  number={3},
  year={2024}
}

@inproceedings{karamcheti2024prismatic,
  title={Prismatic vlms: Investigating the design space of visually-conditioned language models},
  author={Karamcheti, Siddharth and Nair, Suraj and Balakrishna, Ashwin and Liang, Percy and Kollar, Thomas and Sadigh, Dorsa},
  booktitle={Forty-first International Conference on Machine Learning},
  year={2024}
}

@article{fei2025libero,
  title={LIBERO-Plus: In-depth Robustness Analysis of Vision-Language-Action Models},
  author={Fei, Senyu and Wang, Siyin and Shi, Junhao and Dai, Zihao and Cai, Jikun and Qian, Pengfang and Ji, Li and He, Xinzhe and Zhang, Shiduo and Fei, Zhaoye and others},
  journal={arXiv preprint arXiv:2510.13626},
  year={2025}
}

@article{liu2023libero,
  title={Libero: Benchmarking knowledge transfer for lifelong robot learning},
  author={Liu, Bo and Zhu, Yifeng and Gao, Chongkai and Feng, Yihao and Liu, Qiang and Zhu, Yuke and Stone, Peter},
  journal={Advances in Neural Information Processing Systems},
  volume={36},
  pages={44776--44791},
  year={2023}
}

@article{lauri2022partially,
  title={Partially observable markov decision processes in robotics: A survey},
  author={Lauri, Mikko and Hsu, David and Pajarinen, Joni},
  journal={IEEE Transactions on Robotics},
  volume={39},
  number={1},
  pages={21--40},
  year={2022},
  publisher={IEEE}
}

@article{li2024cogact,
  title={Cogact: A foundational vision-language-action model for synergizing cognition and action in robotic manipulation},
  author={Li, Qixiu and Liang, Yaobo and Wang, Zeyu and Luo, Lin and Chen, Xi and Liao, Mozheng and Wei, Fangyun and Deng, Yu and Xu, Sicheng and Zhang, Yizhong and others},
  journal={arXiv preprint arXiv:2411.19650},
  year={2024}
}

@inproceedings{pan2025semantic,
  title={Semantic and sequential alignment for referring video object segmentation},
  author={Pan, Feiyu and Fang, Hao and Li, Fangkai and Xu, Yanyu and Li, Yawei and Benini, Luca and Lu, Xiankai},
  booktitle={Proceedings of the Computer Vision and Pattern Recognition Conference},
  pages={19067--19076},
  year={2025}
}

@article{qian2024streaming,
  title={Streaming long video understanding with large language models},
  author={Qian, Rui and Dong, Xiaoyi and Zhang, Pan and Zang, Yuhang and Ding, Shuangrui and Lin, Dahua and Wang, Jiaqi},
  journal={Advances in Neural Information Processing Systems},
  volume={37},
  pages={119336--119360},
  year={2024}
}

@article{fan2025vlm,
  title={VLM-3R: Vision-Language Models Augmented with Instruction-Aligned 3D Reconstruction},
  author={Fan, Zhiwen and Zhang, Jian and Li, Renjie and Zhang, Junge and Chen, Runjin and Hu, Hezhen and Wang, Kevin and Qu, Huaizhi and Wang, Dilin and Yan, Zhicheng and others},
  journal={arXiv preprint arXiv:2505.20279},
  year={2025}
}

@inproceedings{wang2025continuous,
  title={Continuous 3d perception model with persistent state},
  author={Wang, Qianqian and Zhang, Yifei and Holynski, Aleksander and Efros, Alexei A and Kanazawa, Angjoo},
  booktitle={Proceedings of the Computer Vision and Pattern Recognition Conference},
  pages={10510--10522},
  year={2025}
}

@article{cheang2025gr,
  title={Gr-3 technical report},
  author={Cheang, Chilam and Chen, Sijin and Cui, Zhongren and Hu, Yingdong and Huang, Liqun and Kong, Tao and Li, Hang and Li, Yifeng and Liu, Yuxiao and Ma, Xiao and others},
  journal={arXiv preprint arXiv:2507.15493},
  year={2025}
}

@article{tinyvla,
  title={Tinyvla: Towards fast, data-efficient vision-language-action models for robotic manipulation},
  author={Wen, Junjie and Zhu, Yichen and Li, Jinming and Zhu, Minjie and Tang, Zhibin and Wu, Kun and Xu, Zhiyuan and Liu, Ning and Cheng, Ran and Shen, Chaomin and others},
  journal={IEEE Robotics and Automation Letters},
  year={2025},
  publisher={IEEE}
}

@inproceedings{perez2018film,
  title={Film: Visual reasoning with a general conditioning layer},
  author={Perez, Ethan and Strub, Florian and De Vries, Harm and Dumoulin, Vincent and Courville, Aaron},
  booktitle={Proceedings of the AAAI conference on artificial intelligence},
  volume={32},
  number={1},
  year={2018}
}

@article{jiang2025better,
  title={The better you learn, the smarter you prune: Towards efficient vision-language-action models via differentiable token pruning},
  author={Jiang, Titong and Jiang, Xuefeng and Ma, Yuan and Wen, Xin and Li, Bailin and Zhan, Kun and Jia, Peng and Liu, Yahui and Sun, Sheng and Lang, Xianpeng},
  journal={arXiv preprint arXiv:2509.12594},
  year={2025}
}

@article{song2025reconvla,
  title={Reconvla: Reconstructive vision-language-action model as effective robot perceiver},
  author={Song, Wenxuan and Zhou, Ziyang and Zhao, Han and Chen, Jiayi and Ding, Pengxiang and Yan, Haodong and Huang, Yuxin and Tang, Feilong and Wang, Donglin and Li, Haoang},
  journal={arXiv preprint arXiv:2508.10333},
  year={2025}
}

@article{wang2025vla,
  title={VLA-Adapter: An Effective Paradigm for Tiny-Scale Vision-Language-Action Model},
  author={Wang, Yihao and Ding, Pengxiang and Li, Lingxiao and Cui, Can and Ge, Zirui and Tong, Xinyang and Song, Wenxuan and Zhao, Han and Zhao, Wei and Hou, Pengxu and others},
  journal={arXiv preprint arXiv:2509.09372},
  year={2025}
}

@article{linreasonable,
  title={Reasonable Effectiveness of Random Weighting: A Litmus Test for Multi-Task Learning},
  author={Lin, Baijiong and Ye, Feiyang and Zhang, Yu and Tsang, Ivor},
  journal={Transactions on Machine Learning Research}
}

@article{qian2025geopredict,
  title={GeoPredict: Leveraging Predictive Kinematics and 3D Gaussian Geometry for Precise VLA Manipulation},
  author={Qian, Jingjing and Han, Boyao and Shi, Chen and Xiao, Lei and Yang, Long and Shi, Shaoshuai and Jiang, Li},
  journal={arXiv preprint arXiv:2512.16811},
  year={2025}
}

@article{shi2025memoryvla,
  title={Memoryvla: Perceptual-cognitive memory in vision-language-action models for robotic manipulation},
  author={Shi, Hao and Xie, Bin and Liu, Yingfei and Sun, Lin and Liu, Fengrong and Wang, Tiancai and Zhou, Erjin and Fan, Haoqiang and Zhang, Xiangyu and Huang, Gao},
  journal={arXiv preprint arXiv:2508.19236},
  year={2025}
}

@inproceedings{hong2021vln,
  title={Vln bert: A recurrent vision-and-language bert for navigation},
  author={Hong, Yicong and Wu, Qi and Qi, Yuankai and Rodriguez-Opazo, Cristian and Gould, Stephen},
  booktitle={Proceedings of the IEEE/CVF conference on Computer Vision and Pattern Recognition},
  pages={1643--1653},
  year={2021}
}

@article{zhang2025safevla,
  title={Safevla: Towards safety alignment of vision-language-action model via constrained learning},
  author={Zhang, Borong and Zhang, Yuhao and Ji, Jiaming and Lei, Yingshan and Dai, Josef and Chen, Yuanpei and Yang, Yaodong},
  journal={arXiv preprint arXiv:2503.03480},
  year={2025}
}

@article{ni2021recurrent,
  title={Recurrent model-free rl can be a strong baseline for many pomdps},
  author={Ni, Tianwei and Eysenbach, Benjamin and Salakhutdinov, Ruslan},
  journal={arXiv preprint arXiv:2110.05038},
  year={2021}
}

@inproceedings{hausknecht2015deep,
  title={Deep Recurrent Q-Learning for Partially Observable MDPs.},
  author={Hausknecht, Matthew J and Stone, Peter},
  booktitle={AAAI fall symposia},
  volume={45},
  pages={141},
  year={2015}
}

@article{zhang2024navid,
  title={Navid: Video-based vlm plans the next step for vision-and-language navigation},
  author={Zhang, Jiazhao and Wang, Kunyu and Xu, Rongtao and Zhou, Gengze and Hong, Yicong and Fang, Xiaomeng and Wu, Qi and Zhang, Zhizheng and Wang, He},
  journal={arXiv preprint arXiv:2402.15852},
  year={2024}
}
